\documentclass{article}

\PassOptionsToPackage{numbers, compress}{natbib}
\usepackage[preprint]{neurips_2025}

\usepackage[utf8]{inputenc}
\usepackage[T1]{fontenc}
\usepackage{hyperref}
\usepackage{url}
\usepackage{booktabs}
\usepackage{nicefrac}
\usepackage{microtype}
\usepackage{graphicx}

\usepackage{amsmath}
\usepackage{amssymb}
\usepackage{algorithm}
\usepackage[noend]{algpseudocode}
\usepackage{mathrsfs}
\usepackage{dsfont}
\usepackage{array}
\usepackage{bbm}
\usepackage[mathscr]{eucal}
\usepackage{psfrag}

\usepackage{wasysym}
\usepackage[dvipsnames, table]{xcolor}  
\usepackage{caption}                     
\usepackage{subcaption}
\usepackage{amsthm}
\usepackage{todonotes}
\usepackage{tabulary}
\usepackage{outlines}
\usepackage{thmtools, thm-restate}

\newtheorem{proposition} {Proposition}
\newtheorem{lemma} {Lemma}

\newcommand{\E}{\mathbb{E}}

\definecolor{ian_highlight}{RGB}{100, 2, 2}

\errorcontextlines\maxdimen

\makeatletter
    \newcommand*{\algrule}[1][\algorithmicindent]{\makebox[#1][l]{\hspace*{.5em}\thealgruleextra\vrule height \thealgruleheight depth \thealgruledepth}}%
\newcommand*{\thealgruleextra}{}
\newcommand*{\thealgruleheight}{.75\baselineskip}
\newcommand*{\thealgruledepth}{.25\baselineskip}

\newcount\ALG@printindent@tempcnta
\def\ALG@printindent{%
    \ifnum \theALG@nested>0
        \ifx\ALG@text\ALG@x@notext
        \else
            \unskip
            \addvspace{-1pt}
            \ALG@printindent@tempcnta=1
            \loop
                \algrule[\csname ALG@ind@\the\ALG@printindent@tempcnta\endcsname]%
                \advance \ALG@printindent@tempcnta 1
            \ifnum \ALG@printindent@tempcnta<\numexpr\theALG@nested+1\relax
            \repeat
        \fi
    \fi
    }%
\usepackage{etoolbox}
\makeatother

\newbox\statebox
\newcommand{\myState}[1]{%
    \setbox\statebox=\vbox{#1}%
    \edef\thealgruleheight{\dimexpr \the\ht\statebox+1pt\relax}%
    \edef\thealgruledepth{\dimexpr \the\dp\statebox+1pt\relax}%
    \ifdim\thealgruleheight<.75\baselineskip
        \def\thealgruleheight{\dimexpr .75\baselineskip+1pt\relax}%
    \fi
    \ifdim\thealgruledepth<.25\baselineskip
        \def\thealgruledepth{\dimexpr .25\baselineskip+1pt\relax}%
    \fi
    \State #1%
    \def\thealgruleheight{\dimexpr .75\baselineskip+1pt\relax}%
    \def\thealgruledepth{\dimexpr .25\baselineskip+1pt\relax}%
}

\newcommand{\CE}{\mathrm{CE}}



\title{Planning to Learn}

\author{
  Ian Osband \\
  Google DeepMind \\
  \texttt{iosband@google.com}
}

\begin{document}

\maketitle

\begin{abstract}
Policy-gradient methods are central to modern reinforcement learning, including LLM post-training.
When they struggle, the usual suspects are exploration, credit assignment and action-sampling noise.
Classification has none of them.
A classifier is a policy whose expected reward, its \emph{expected accuracy}, is the probability it assigns to the correct label, and because that label is known, the policy gradient is exact and smooth.
Yet exact policy gradient loses to cross-entropy, even on expected accuracy.
The exact gradient is myopic: it values an update only by what it buys now, but each update also sets where the next one starts, so an update's value depends on how much learning remains.
Viewed this way, cross-entropy is \emph{patient accuracy}, the total error an example would pay if its log-odds rose at unit speed forever, while exact policy gradient is the zero-horizon limit.
Truncating this total at the learning that remains yields the \emph{horizon loss}, a one-line change that moves from cross-entropy toward exact policy gradient as training runs out.
In a simple allocation model, it provably escapes the trap that catches each endpoint.
On MNIST and on ImageNet with ResNet-50, ResNet-101 and ViT-S/16, the horizon loss improves top-1 accuracy over cross-entropy at a flat learning rate, and the gain grows with label noise.
\end{abstract}

\section{Introduction}
\label{sec:introduction}

\begin{figure}[ht!]
\centering
\includegraphics[width=0.85\textwidth]{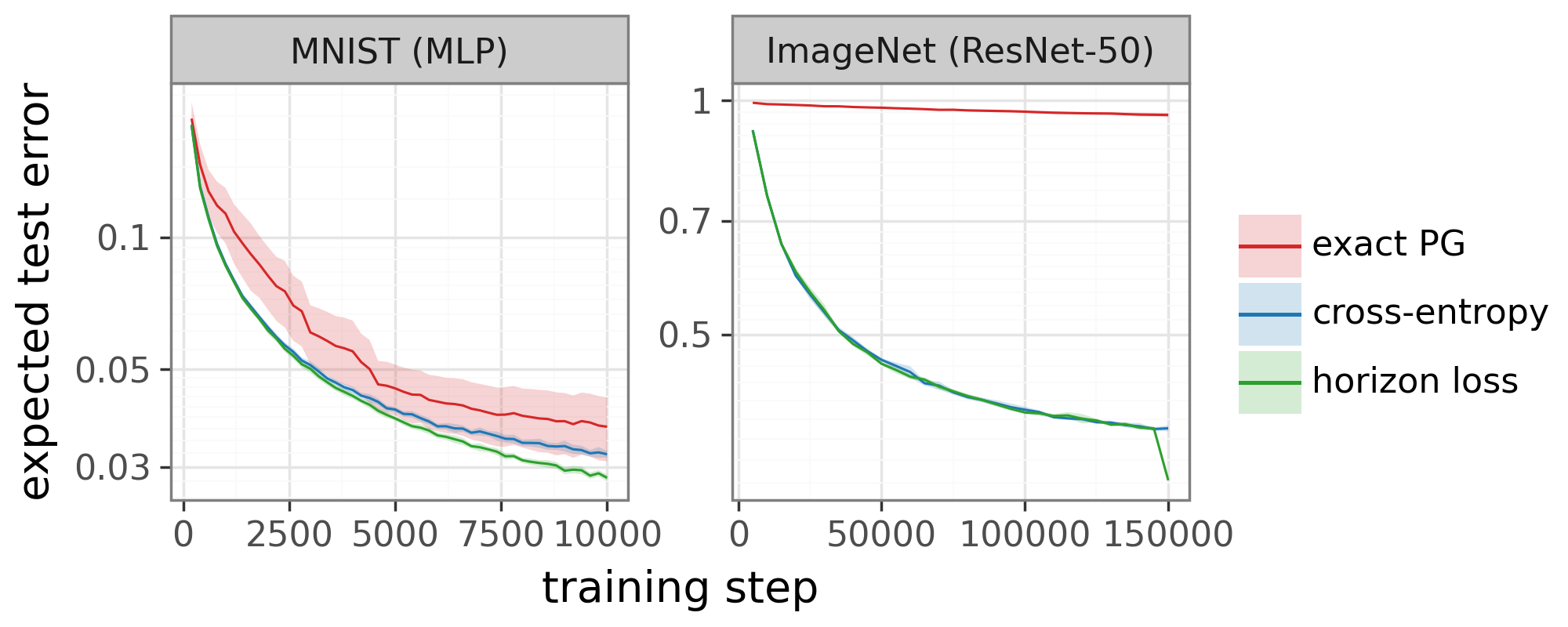}
\caption{Expected test error (log scale) over training at a shared flat learning rate, MNIST (MLP) $\vert$ ImageNet (ResNet-50). Exact policy gradient, computed by summing over labels, falls far behind cross-entropy even on expected test accuracy. The horizon loss (Section~\ref{sec:horizon}) pulls ahead of cross-entropy late in training.}
\label{fig:curves}
\end{figure}

Reinforcement learning has become a central stage of training frontier language models, whose post-training for reasoning runs on policy-gradient updates~\citep{openai2024o1,guo2025deepseek,shao2024deepseekmath}.
It is also harder to get right than supervised pretraining, and when it struggles, the usual suspects are what reinforcement learning adds: exploration, credit assignment and action-sampling noise~\citep{sutton2018reinforcement}.
Classification has none of them.
A classifier is a policy that samples a label and earns reward one when it is correct, so its expected reward is its \emph{expected accuracy}, the same expected-reward form used in policy-gradient post-training.
Every example is observed, the reward is immediate, and every label can be summed over, so the policy gradient of expected accuracy can be computed exactly~\citep{williams1992simple,sutton1999policy}.
Yet following the exact gradient of expected accuracy loses to cross-entropy, even on expected accuracy (Figure~\ref{fig:curves}).
On ImageNet with a ResNet-50 at a shared learning rate, exact policy gradient reaches 4.2\% expected accuracy and cross-entropy reaches 62\%.
The policy gradient struggles here with every usual reinforcement-learning difficulty removed.

Why does the exact gradient of accuracy lose to a surrogate?
A gradient answers a question about one update: which infinitesimal change most improves expected accuracy now.
Training makes many updates, and each one changes the state from which the next is made.
Consider two examples whose correct labels have probability $0.9$ and $0.01$, and suppose learning can raise the log-odds of either at the same cost.
The gradient favors the first, because a small step in log-odds raises its probability about nine times as much ($0.9 \times 0.1 = 0.09$ against $0.01 \times 0.99 \approx 0.01$).
With little training left, that is the right call, since only the first can gain much in time.
With plenty left, the budget is better spent on the second, which gains almost nothing at first but has far more to gain before training ends.
The best update depends on how much training remains, and the gradient cannot see it (Section~\ref{sec:allocation}).

Our central proposal is to treat learning itself as a sequential resource-allocation problem.
Every loss divides learning across examples, and in doing so trades accuracy now against accuracy later, much as a reinforcement-learning agent trades immediate against future reward.
Exact policy gradient weights each example by $p(1-p)$, the rate at which its accuracy improves now.
Up to positive scale, this is also the negative gradient of categorical mean absolute error, which \citet{zhang2018generalized} found trains poorly.
Cross-entropy weights each example by $1-p$, the accuracy it could still gain with unlimited learning.
This weight comes from \emph{patient accuracy}: cross-entropy equals the total error an example would pay if all future learning went to it, raising its log-odds at unit speed (Section~\ref{sec:patient}).
The allocation model explains how this patience helps, and when it wastes budget.
Cross-entropy spends most on the examples furthest from success, including those the remaining run cannot fix.
The \emph{horizon loss} weights each example by the accuracy it could gain with only the learning that remains, which we estimate as the sum of the learning rates still to come, times a calibration constant $\kappa$.
With a long horizon it approaches cross-entropy, and as the budget runs out, exact policy gradient.

This view makes three predictions.
The horizon loss should help most late in training, when little budget remains.
Its gain should grow with the number of examples that cannot be fixed in time.
And the gain should require a horizon that shrinks over training.
On MNIST and on ImageNet with ResNet-50, ResNet-101 and ViT-S/16, the horizon loss improves top-1 accuracy over cross-entropy at a flat learning rate, by 1.0--2.4 points on ImageNet, and the gain arrives late (Section~\ref{sec:both}).
The gain grows with label noise (Section~\ref{sec:noise}), and constant, growing and fixed horizons do not reproduce it (Section~\ref{sec:shape}).
Planning cross-entropy the same way lowers ImageNet test NLL from 1.467 to 1.364 (Section~\ref{sec:hxent}).

\section{Learning as Allocation}
\label{sec:allocation}

A classifier with logits $f_1, \ldots, f_K$ defines a policy: sample a label from the softmax and earn reward one if it is correct.
For an example with label $y$, the probability of sampling the correct label depends on the logits only through the \emph{logit gap}
\begin{equation}
    z = f_y - \log \sum_{k \ne y} e^{f_k},
    \label{eq:gap}
\end{equation}
since $p = \sigma(z) = 1/(1+e^{-z})$ for any number of classes.
Expected accuracy is $p$, the expected reward of the policy.
Top-1 accuracy instead asks whether $f_y > \max_{k \ne y} f_k$.
With a known label, the sampled action can be marginalized, so apart from minibatch sampling the policy gradient is exact~\citep{williams1992simple}.
It sits beside the negative gradient of cross-entropy $-\log p$:
\begin{equation}
    \nabla_\theta p = p(1-p)\,\nabla_\theta z,
    \qquad
    -\nabla_\theta \CE = (1-p)\,\nabla_\theta z.
    \label{eq:weights}
\end{equation}
Both updates push each example in the same direction and differ only in how hard: exact policy gradient weights it by $p(1-p)$, cross-entropy by $1-p$.
Classification theory motivates cross-entropy as a smooth convex surrogate for the discontinuous top-1 decision~\citep{zhang2004statistical,bartlett2006convexity}.
Expected accuracy needs no surrogate, since it is already smooth, and neither loss is convex in the network parameters.
Whatever cross-entropy gains here does not come from smoothness.

Training an ImageNet classifier for 30 epochs is a typical instance of the problem we study.
The data are fixed, the number of updates is fixed, and what counts is accuracy when training stops.
Formally, a fixed optimizer and learning-rate schedule train a network on a fixed labeled dataset for $T$ steps, and the only choice is the loss $\ell$, which may change with the training step.
Writing $p_T(x,y)$ for the probability that the final network samples the correct label $y$ on input $x$, the goal is
\begin{equation}
    \max_{\ell}\; \mathbb{E}_{(x,y)}\big[p_T(x,y)\big].
    \label{eq:problem}
\end{equation}
A loss is judged by expected accuracy after $T$ updates rather than at its own minimizer, so what matters is how it spreads those updates across examples.

With many examples, the weights in~\eqref{eq:weights} decide where those updates go.
To isolate that allocation problem, we strip away the network and let the learner move each gap directly.
Take $N$ independent examples with gaps $z = (z_1, \ldots, z_N)$ and score a state by its total expected accuracy $V(z) = \sum_i \sigma(z_i)$.
The learner has $H$ units of logit progress to divide among them, standing in for the training that remains.
The best achievable final accuracy is
\begin{equation}
    V_H^\star(z) = \max_{u \ge 0,\, \|u\|_1 \le H} V(z+u),
    \label{eq:oracle}
\end{equation}
and $\Delta_H^\star(z) = V_H^\star(z) - V(z)$ is the gain still available.
A weighting rule $w$ spends the budget continuously, in proportion to its weights:
\begin{equation}
    \dot z_i(\tau) = \frac{w(z_i(\tau), h_\tau)}{\sum_j w(z_j(\tau), h_\tau)},
    \qquad h_\tau = H - \tau,
    \qquad 0 \le \tau \le H.
    \label{eq:flow}
\end{equation}
Here $\tau$ is budget spent and $h_\tau$ is budget remaining.
Normalizing the total rate to one removes the step size, so only the split matters.
Reading off~\eqref{eq:weights}, exact policy gradient uses $w_{\mathrm{PG}}(z) = \sigma'(z)$ and cross-entropy uses $w_{\CE}(z) = 1-\sigma(z)$.
With orthonormal example gradients, normalized gradient ascent on expected accuracy produces exactly this flow.

\begin{restatable}[Exact policy gradient spends by immediate gain]{lemma}{greedy}
\label{lem:greedy}
If the example gradients $\{\nabla_\theta z_i\}_{i=1}^N$ are orthonormal, an infinitesimal step of exact policy gradient moves each gap in proportion to $\sigma'(z_i)$.
After rescaling so that $\sum_i \dot z_i = 1$, it is the flow~\eqref{eq:flow} with $w_{\mathrm{PG}}$.
\end{restatable}

\begin{figure}[th!]
\centering
\begin{subfigure}{0.32\textwidth}\includegraphics[width=\linewidth]{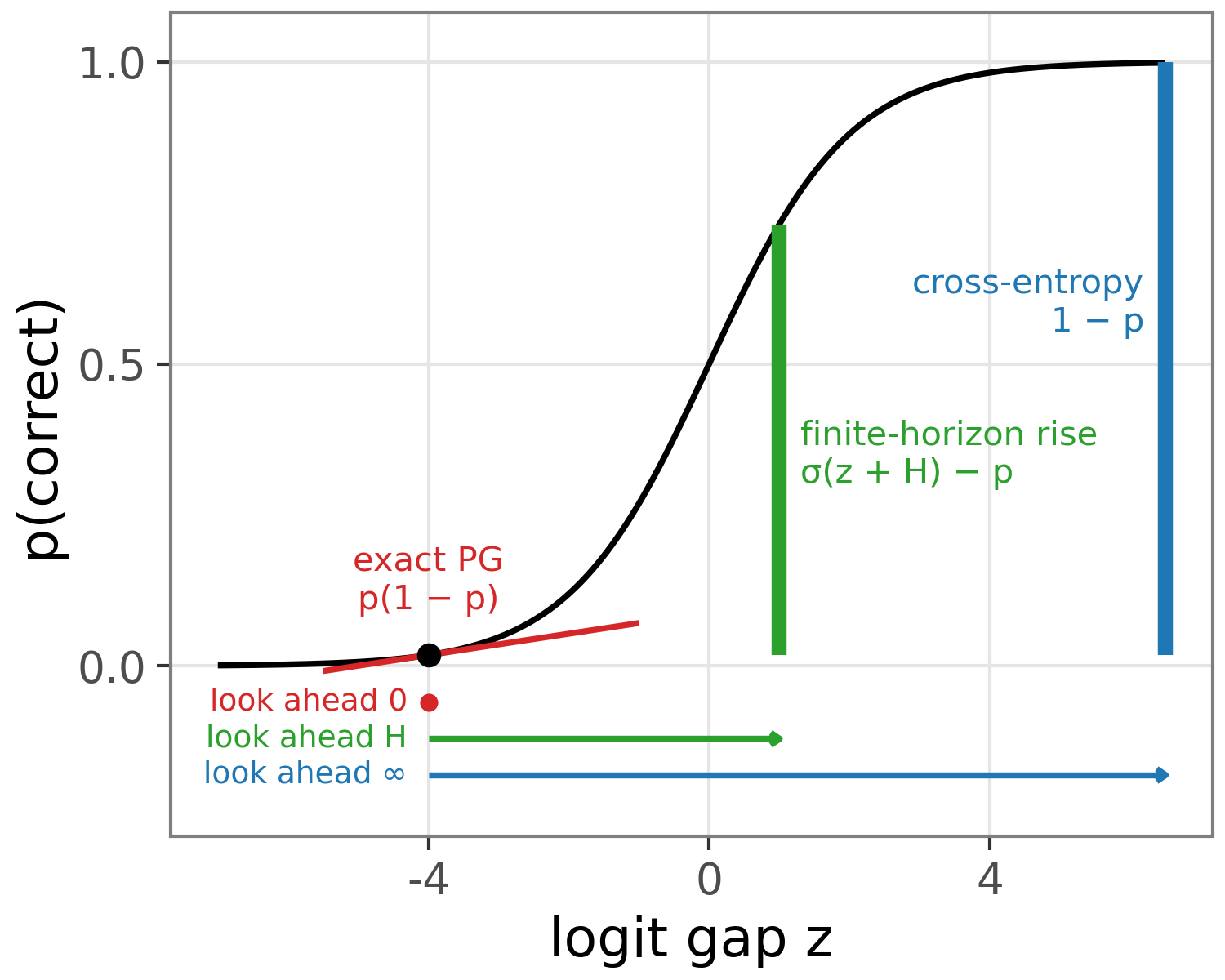}\caption{}\end{subfigure}\hfill
\begin{subfigure}{0.32\textwidth}\includegraphics[width=\linewidth]{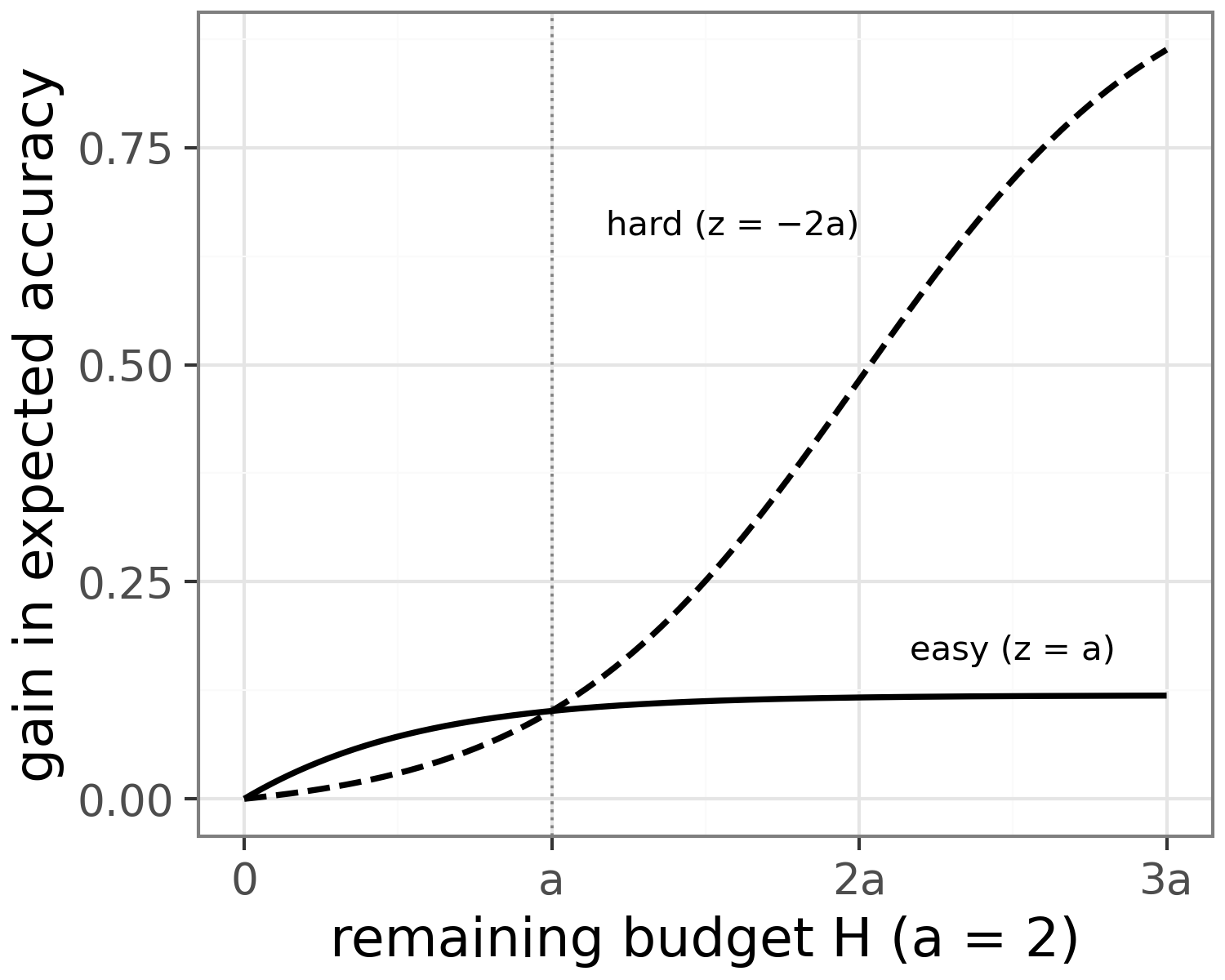}\caption{}\end{subfigure}\hfill
\begin{subfigure}{0.32\textwidth}\includegraphics[width=\linewidth]{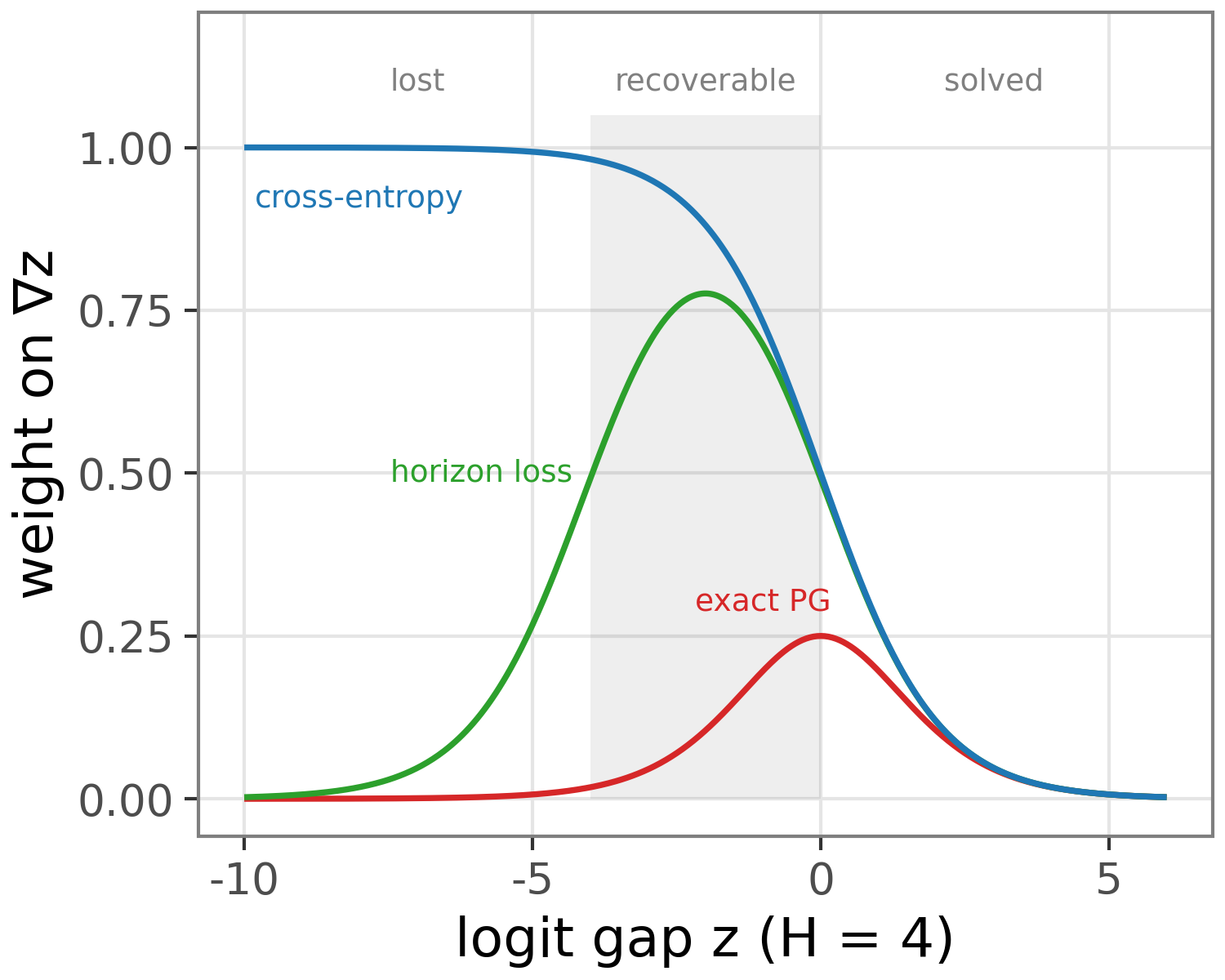}\caption{}\end{subfigure}
\caption{(a) Look-ahead from an example at $z=-4$. Exact policy gradient uses the local slope $p(1-p)$, the finite horizon uses the rise $\sigma(z+H)-p$ over the next $H$ units of learning, and cross-entropy uses the full remaining rise $1-p$. (b) Gain in expected accuracy from spending the whole budget $H$ on each example of Proposition~\ref{prop:reversal} ($a=2$); the better example switches at $H=a$. (c) Weight on $\nabla z$ against the gap at $H=4$. Exact policy gradient concentrates near even odds, cross-entropy gives nearly maximal weight to hopeless examples, and the horizon loss targets examples still within reach.}
\label{fig:theory}
\end{figure}

The weight $\sigma'(z)$ is the rate at which accuracy rises at the current gap, so exact policy gradient favors examples near even odds however much budget remains, and that ranking can be wrong.
Take an easy example at gap $a$ and a hard one at gap $-2a$.
A short budget should go to the easy example, which sits on the steep part of the sigmoid.
A longer budget should go to the hard one, which gains little at first but can be carried to even odds, where it gains more than the easy example can (Figure~\ref{fig:theory}b).

\begin{restatable}[The best allocation reverses with the budget]{proposition}{reversal}
\label{prop:reversal}
Consider two examples at gaps $(a, -2a)$ with $a > 0$ and budget $H$.
If $H < a$, the optimal allocation spends all of $H$ on the easy example.
If $a < H \le 3a$, it spends all of $H$ on the hard example.
\end{restatable}

Exact policy gradient cannot make this reversal.
Its weights satisfy $\sigma'(a) > \sigma'(-2a)$ and contain no $H$, so it initially gives more weight to the easy example whatever the budget, and multiplying every weight by a common factor changes nothing.
At $H = 2a$ it captures only a fraction $4\sigma'(a)$ of the available gain, which vanishes exponentially with $a$ (Section~\ref{sec:traps}).

\section{Cross-Entropy Is Patient Accuracy}
\label{sec:patient}

Suppose all remaining learning went to one example.
Its gap would then rise at unit speed, $z \to z+s$, and a learner that cares about accuracy would pay error $1-\sigma(z+s)$ along the way.
Summed over the next $H$ units of learning, this \emph{future error} is
\begin{equation}
    C_H(z) = \int_0^H [1-\sigma(z+s)]\,ds = \CE(z) - \CE(z+H),
    \label{eq:future}
\end{equation}
because $\tfrac{d}{dz}\CE(z) = -(1-\sigma(z))$.
As $H \to \infty$ the last term vanishes, so
\[
    \CE(z) = \int_0^\infty [1-\sigma(z+s)]\,ds.
\]
Along this unit-speed log-odds path, cross-entropy is the total error an example would pay if training never stopped.
This is the sense in which it is patient accuracy.
For a single Bernoulli decision, the path is exactly the natural policy-gradient flow, since the Fisher information $p(1-p)$ equals $\partial_z p$~\citep{kakade2001natural}.
The path is counterfactual and need not match the trajectory of a shared-parameter optimizer.
Cross-entropy also prices each example as if the budget were its alone, a tractable surrogate for the shared-budget problem~\eqref{eq:oracle}, not its solution.

\subsection{The Learning Horizon}
\label{sec:horizon}

A finite horizon changes the weight.
The negative gradient of $C_H$ is $\sigma(z+H) - \sigma(z)$, the accuracy the example would gain from the whole remaining budget, and it factors exactly:
\begin{equation}
    \frac{\sigma(z+H) - \sigma(z)}{1-e^{-H}} = \sigma(z+H)\,(1-\sigma(z)).
    \label{eq:secant}
\end{equation}
The \emph{horizon loss}
\begin{equation}
    L_H(z) = \frac{C_H(z)}{1-e^{-H}}
    \label{eq:hloss}
\end{equation}
therefore weights $\nabla z$ by the remaining error times the chance of success within the horizon.
At $H = 0$ this weight is $\sigma'(z)$, exact policy gradient, and as $H \to \infty$ it is $1-\sigma(z)$, cross-entropy (Proposition~\ref{prop:endpoints}).
Both endpoint weights in~\eqref{eq:weights} have this form.
They share the factor $1-p$, the error still left on the example, and multiply it by a chance of success: $p$ now for exact policy gradient, and $1 = \lim_{h \to \infty} \sigma(z+h)$ after unlimited learning for cross-entropy (Figure~\ref{fig:theory}a).
Exact policy gradient looks zero distance ahead and cross-entropy looks infinitely far, but neither looks at the budget that actually remains.
For a learner with $h$ units left, the \emph{horizon rule} allocates by attainable gain rather than by slope:
\begin{equation}
    w_{\mathrm{hor}}(z,h) = \sigma(z+h)\,(1-\sigma(z)).
    \label{eq:whor}
\end{equation}
The normalization by $1-e^{-H}$ is a common factor at fixed $H$, so it leaves the allocation unchanged and keeps the scale continuous between the two endpoints.

In a neural network, the remaining logit progress is not known in advance.
We approximate it by the learning rate still to come,
\begin{equation}
    H_t = \kappa \sum_{t \le t' < T} \eta_{t'},
    \label{eq:horizon}
\end{equation}
where $\kappa$ converts step size into logit progress and is calibrated once per dataset and schedule (Appendix~\ref{app:details}).
With a flat learning rate, $H_t$ falls linearly to zero.
When the initial horizon is large relative to the gaps, the horizon loss starts close to cross-entropy, since $\sigma(z+H_t) \approx 1$ for every example, and it moves toward exact policy gradient as training runs out.

The same construction applies to any per-example loss $\ell$.
The planning operator
\begin{equation}
    P_H[\ell](z) = \int_0^H \ell(z+s)\,ds
    \label{eq:planning}
\end{equation}
sums the loss along the same path.
Future error is $C_H = P_H[1-\sigma]$, and applying the operator to cross-entropy gives the horizon cross-entropy of Section~\ref{sec:hxent}.
Every such loss is a separable plan: it prices each example as if the whole remaining budget were available to it alone.
The horizon loss escapes both traps below without coupling examples, and Appendix~\ref{app:planning} shows where any separable loss must fail.

\subsection{Two Wrong Horizons}
\label{sec:traps}

Each endpoint wastes budget in its own way.
Exact policy gradient cannot tell which side of even odds an example is on.
Since $\sigma'(a) = \sigma'(-a)$, it polishes an example already likely right as eagerly as it rescues one likely wrong.
Cross-entropy cannot tell whether an example is within reach.
Its weight $1-\sigma(z)$ is largest on the examples with the least chance of success, so it spends most on lost causes that the remaining budget cannot fix.
Two small problems make each failure exact.

The \emph{polishing trap} starts from $(a, -2a)$ with $H = 2a$, where Proposition~\ref{prop:reversal} says to spend everything on the hard example.
The \emph{lost-cause trap} starts from one example at even odds and $M$ lost causes at gap $-a$, with $0 < H < a$.
The lost causes cannot reach even odds within the budget, so the optimum spends everything on the first example and gains $\Delta_H^\star = \sigma(H) - \tfrac12$.
For a rule $A$ with weight $w_A$, let $z^A(\tau)$ follow the flow~\eqref{eq:flow} from the initial state $z$, and let
\begin{equation}
    J_A = V\big(z^A(H)\big)
    \label{eq:JA}
\end{equation}
be the final total expected accuracy, with gain $\Delta_A = J_A - V(z)$.
We measure failure by the share of available gain captured, $\Delta_A / \Delta_H^\star$, and by the regret $V_H^\star - J_A$.

\begin{proposition}[Each endpoint falls into one trap]
\label{prop:traps}
(i) On the polishing trap, exact policy gradient ends at $(2a, -a)$ and captures exactly
\[
    \frac{\Delta_{\mathrm{PG}}}{\Delta_H^\star} = 4\sigma'(a)
\]
of the optimal gain, which vanishes exponentially in $a$.
(ii) On the lost-cause trap, cross-entropy gains at most
\[
    \Delta_{\CE} \le \frac{H}{4M+4} + He^{H-a};
\]
for fixed $H$, its share of the optimal gain tends to zero as $M \to \infty$ and $a - H \to \infty$.
\end{proposition}

Exact policy gradient finishes polishing the easy example and leaves the hard one short of even odds.
Even the greediest allocator, spending each increment on the steepest slope, ends at the same point (Appendix~\ref{app:greed}).
Cross-entropy weights every lost cause at least as heavily as the one example it can finish, so that example receives at most a $1/(M+1)$ share of the budget.

The horizon loss escapes both traps, with regret that shrinks exponentially as the distractions move outside the horizon and grows only linearly with their number.

\begin{proposition}[The horizon loss escapes both traps]
\label{prop:horizon}
Under the flow~\eqref{eq:flow} with weight $w_{\mathrm{hor}}$~\eqref{eq:whor} and $J_{\mathrm{hor}}$ as in~\eqref{eq:JA}, the regret on the polishing trap (i) and the lost-cause trap (ii) satisfies
\[
    \text{(i)}\;\; V_H^\star - J_{\mathrm{hor}} \le 2ae^{-a},
    \qquad
    \text{(ii)}\;\; V_H^\star - J_{\mathrm{hor}} \le HMe^{H-a}.
\]
The latter is at most $\epsilon$ whenever $a \ge H + \log(HM/\epsilon)$.
\end{proposition}

Both bounds have the same source.
The horizon weight $\sigma(z+h)(1-\sigma(z))$ vanishes at both extremes: a solved example has little error left, and a lost example has little chance of success within the remaining horizon.
An example falls short of the value it would reach with the whole budget by at most the total weight spent elsewhere (Lemma~\ref{lem:leakage}).
That distracting weight is at most $e^{-a}$ on the polishing trap and $Me^{H-a}$ on the lost-cause trap, and integrating over the budget gives Proposition~\ref{prop:horizon}.
Appendix~\ref{app:proofs} proves each result in full.

\section{Experiments}
\label{sec:experiments}

The allocation model has independent examples and a fixed budget of logit progress.
A network shares parameters across examples and trains with Adam or AdamW, so these experiments ask whether the allocation principle remains useful outside the model.
On MNIST we train a two-hidden-layer MLP over 30 seeds, and on ImageNet we train ResNet-50, ResNet-101 and ViT-S/16 for 30 epochs over 3--10 seeds (Appendix~\ref{app:details}).
Within each recipe only the loss changes: exact policy gradient, cross-entropy or the horizon loss.
We use a flat learning rate by default, so the remaining budget $H_t$ in~\eqref{eq:horizon} falls linearly and the loss is the only part of the recipe that accounts for how much training remains.
Section~\ref{sec:both} also reports cosine decay.

The horizon loss introduces one new parameter, the calibration $\kappa$ in~\eqref{eq:horizon}.
We use one value per dataset and schedule ($\kappa = 1$ on MNIST and $3$ on ImageNet at a flat learning rate, and $0.3$ under cosine decay on both).
We choose the ImageNet values on ResNet-50 and reuse them for every architecture and noise level.
At a flat learning rate every method shares one learning rate, and $\kappa$ is chosen on test accuracy with no separate validation split (Appendix~\ref{app:details}), so each comparison holds within one training recipe.
At the default $\kappa = 1$, the horizon loss still improves expected accuracy over cross-entropy on every ImageNet model and schedule we ran, and top-1 accuracy on all but ViT-S/16 at a flat learning rate (Appendix~\ref{app:robustness}).

We report both expected and top-1 accuracy.
Expected accuracy $\E[p_y]$ is the objective of the theory, but it can improve through sharper probabilities without changing a prediction (Proposition~\ref{prop:calibration}).
Top-1 accuracy is invariant to temperature rescaling, so we lead with it.

\subsection{Planning Beats Both}
\label{sec:both}

The horizon loss contains exact policy gradient as its zero-horizon endpoint, but does not inherit its failure.
At the shared learning rate of $10^{-3}$, ResNet-50 expected accuracy rises from 4.2\% under exact policy gradient to 67.5\% under the horizon loss.
On MNIST, where exact policy gradient does train, the horizon loss still cuts its expected error by 24\% and its top-1 error by 29\%.

\begin{figure}[ht!]
\centering
\includegraphics[width=\textwidth]{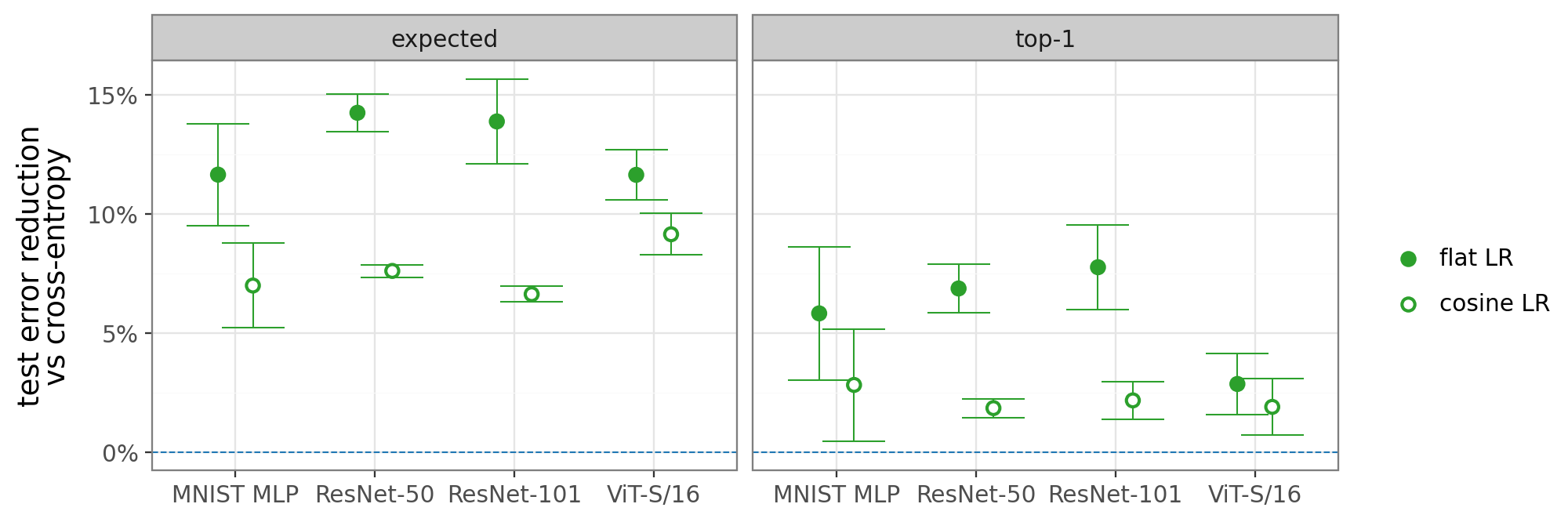}
\caption{Relative reduction in test error over cross-entropy for expected error (left) and top-1 error (right), mean $\pm 2$ SEM over seeds, with the SEM of the ratio propagated from the two independent means. Filled: flat learning rate; open: cosine. The horizon loss has lower mean error than cross-entropy for every model and schedule shown; cosine decay shrinks the gain.}
\label{fig:summary}
\end{figure}

The harder comparison is cross-entropy.
At a flat learning rate, the horizon loss lowers top-1 error on all four models (Figure~\ref{fig:summary}): by 1.0--2.4 points on ImageNet and $0.16 \pm 0.04$ on MNIST.
Expected accuracy improves more, by 4.7--5.4 points on ImageNet, partly by sharpening (Proposition~\ref{prop:calibration}).

Cosine decay shrinks the gain to about 0.6 points of top-1 accuracy on ImageNet, though the horizon loss still improves on cross-entropy on every model.
These comparisons do not separate example allocation from effective step-size effects late in training.
A hand-built horizon, cross-entropy followed by a late switch to exact policy gradient, roughly matches the horizon loss on MNIST and slightly exceeds it under cosine decay on ResNet-50 (Appendix~\ref{app:robustness}), so much of the gain comes from moving toward exact policy gradient late in training, not from the exact form of the horizon weight.

On ImageNet the gain arrives late (Figure~\ref{fig:curves}).
On ResNet-50, a flat learning rate of $10^{-3}$ over 150k steps gives $H_0 \approx 150\kappa = 450$, far beyond the scale of the gaps still in play.
For most of training $\sigma(z+H_t) \approx 1$ on those examples, and the horizon loss is nearly cross-entropy.
The two separate only when $H_t$ falls to the scale of the gaps, in the last few percent of training, and that is where the error curves diverge.
On MNIST, $H_0 \approx 10$ is already on the scale of the gaps, and the curves separate much earlier.

\subsection{The Gain Grows with Label Noise}
\label{sec:noise}

\begin{figure}[ht!]
\centering
\begin{subfigure}{0.32\textwidth}\includegraphics[width=\linewidth]{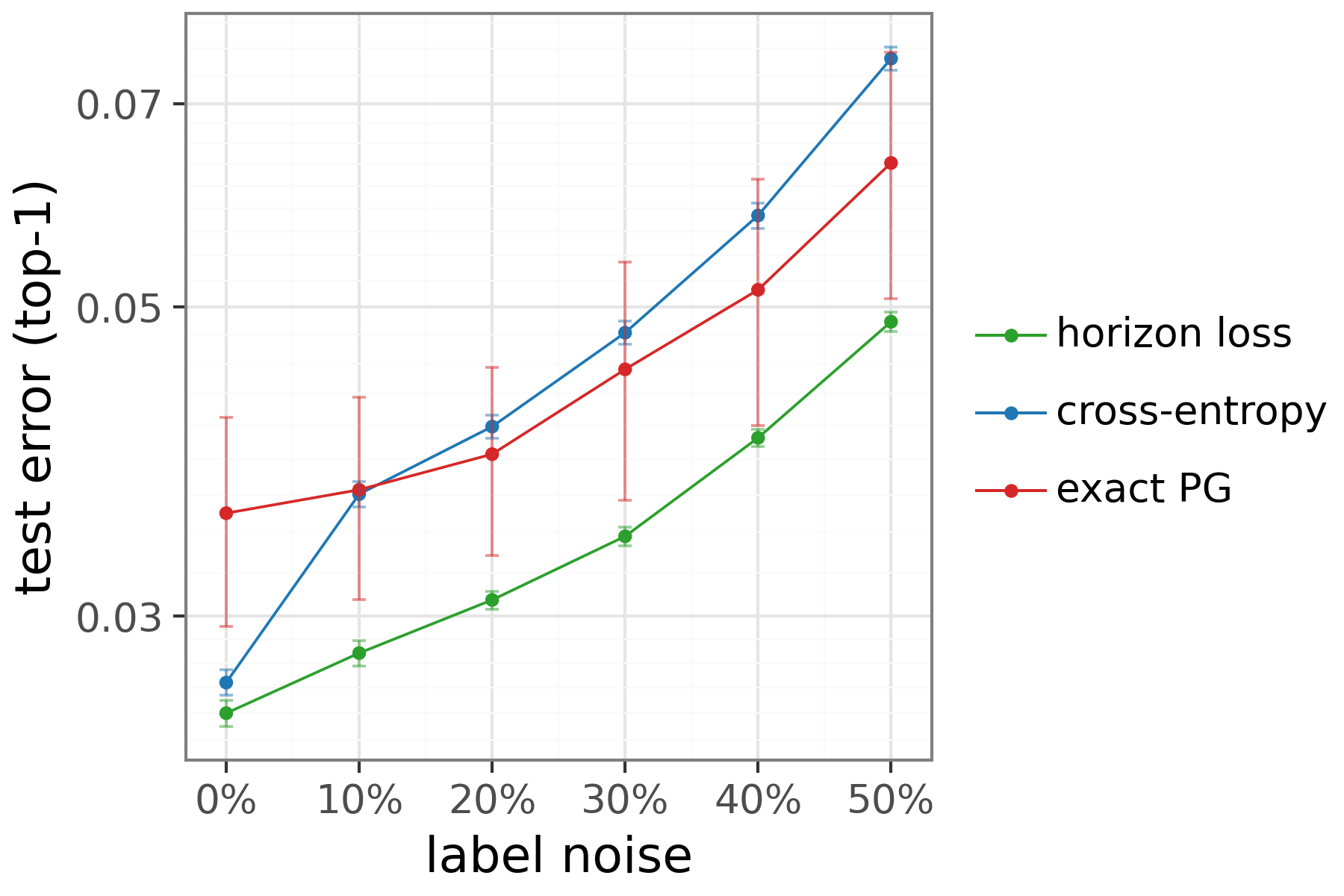}\caption{MNIST}\end{subfigure}\hfill
\begin{subfigure}{0.32\textwidth}\includegraphics[width=\linewidth]{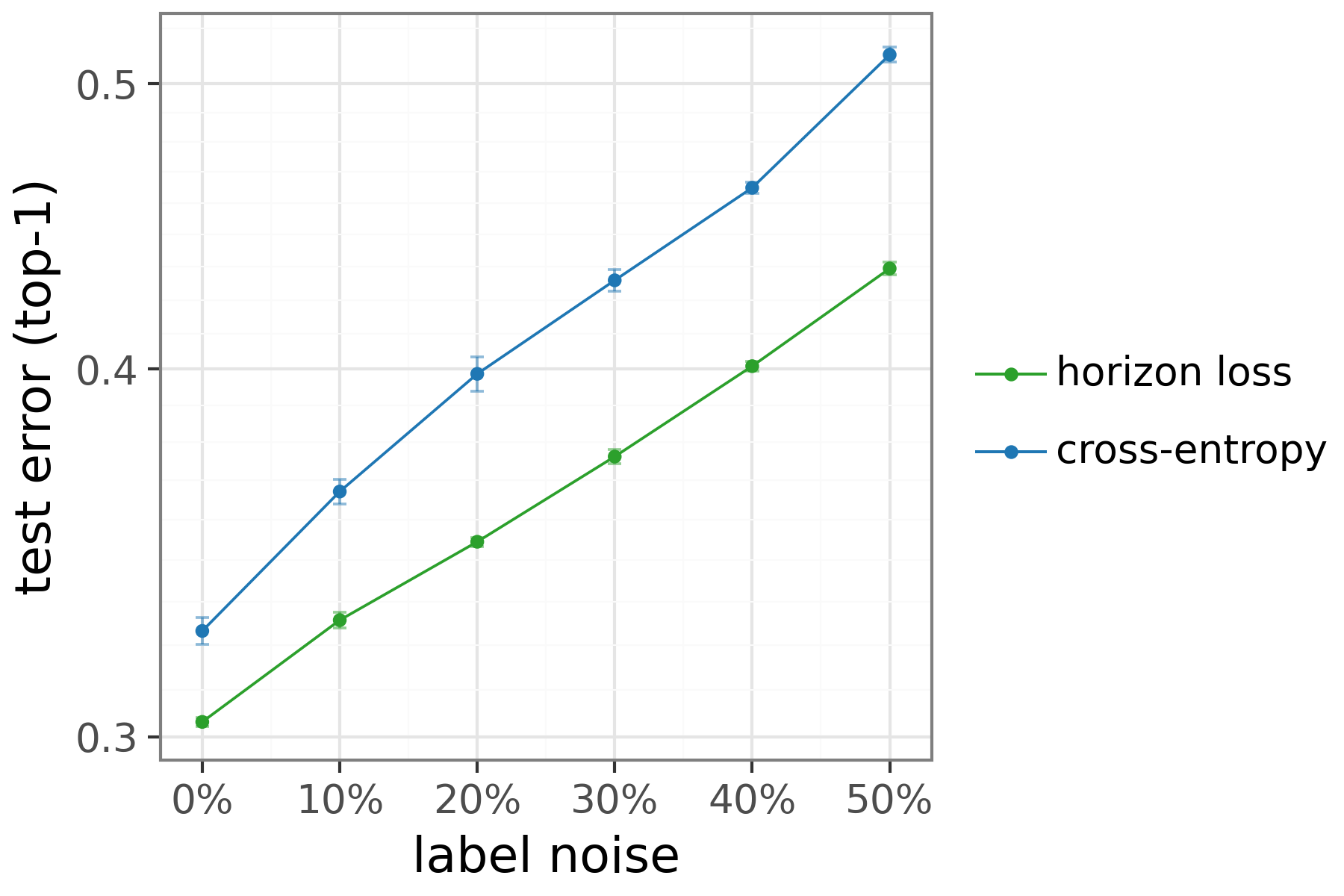}\caption{ResNet-50}\end{subfigure}\hfill
\begin{subfigure}{0.32\textwidth}\includegraphics[width=\linewidth]{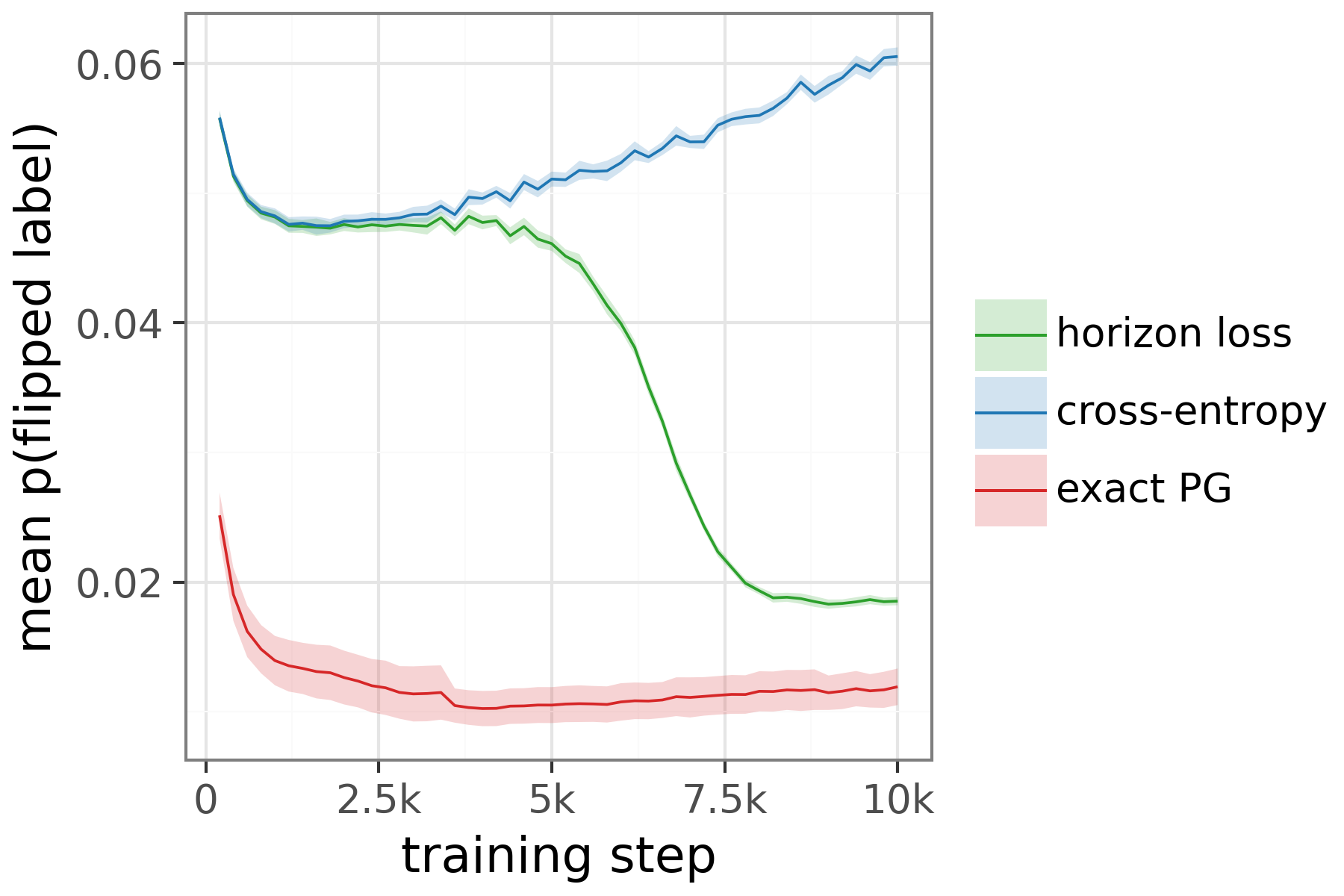}\caption{MNIST, 30\% noise}\end{subfigure}
\caption{(a, b) Top-1 test error against the fraction of flipped training labels, mean $\pm 2$ SEM over 30 seeds on MNIST and 3 seeds per noise level on ResNet-50 (10 at 0\%), for one fixed corruption of the labels. ResNet-50 has no exact-PG noise sweep. (c) Mean probability assigned to the flipped label on mislabeled training examples over training. The noise sweep is separate from the main runs of Section~\ref{sec:both}.}
\label{fig:noise}
\end{figure}

Flipped labels create examples that can consume training while hurting clean-test accuracy.
Once the network learns the true class of a mislabeled training example, the corrupted label can have low probability.
As the horizon shrinks, such examples can fall outside it and become lost causes in the sense of Section~\ref{sec:traps}.
Cross-entropy should keep spending on them while the horizon loss lets them go, so the gain over cross-entropy should grow with the noise rate.
We flip a fraction of training labels uniformly at random and keep test labels clean (Appendix~\ref{app:details}).
The top-1 gain rises monotonically with noise, from 3.3 points at 10\% noise to 7.3 at 50\% on ResNet-50 and from 0.1 to 2.7 points on MNIST (Figure~\ref{fig:noise}a,b).

Figure~\ref{fig:noise}c shows the probability each model assigns to the corrupted label on mislabeled training examples, on MNIST at 30\% noise.
Under cross-entropy it dips early, then rises for the rest of training as the network memorizes the flipped labels.
The horizon loss follows cross-entropy for the first half of training, then drives it down to about a third of cross-entropy's final value.
Gaps never fall in the allocation model, so this decline runs through shared parameters, not the per-example flow.

The endpoint comparison also reverses.
On MNIST, exact policy gradient has higher top-1 error than cross-entropy without noise, but lower mean top-1 error from 20\% noise on (Figure~\ref{fig:noise}a).
As lost causes multiply, ignoring them starts to pay (Proposition~\ref{prop:traps}(ii)).
The horizon loss beats both at every noise level.

\subsection{Long Early, Short Late}
\label{sec:shape}

\begin{figure}[ht!]
\centering
\begin{subfigure}{0.49\textwidth}\includegraphics[width=\linewidth]{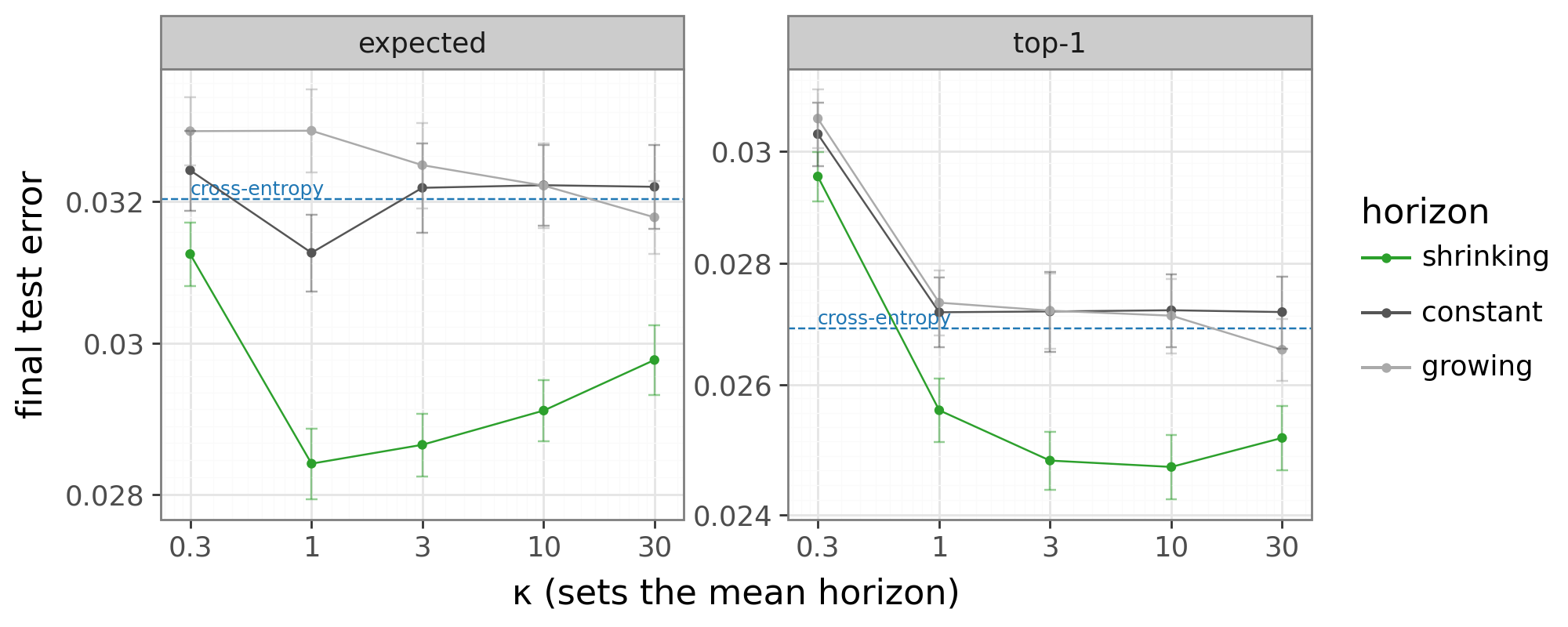}\caption{MNIST}\end{subfigure}\hfill
\begin{subfigure}{0.49\textwidth}\includegraphics[width=\linewidth]{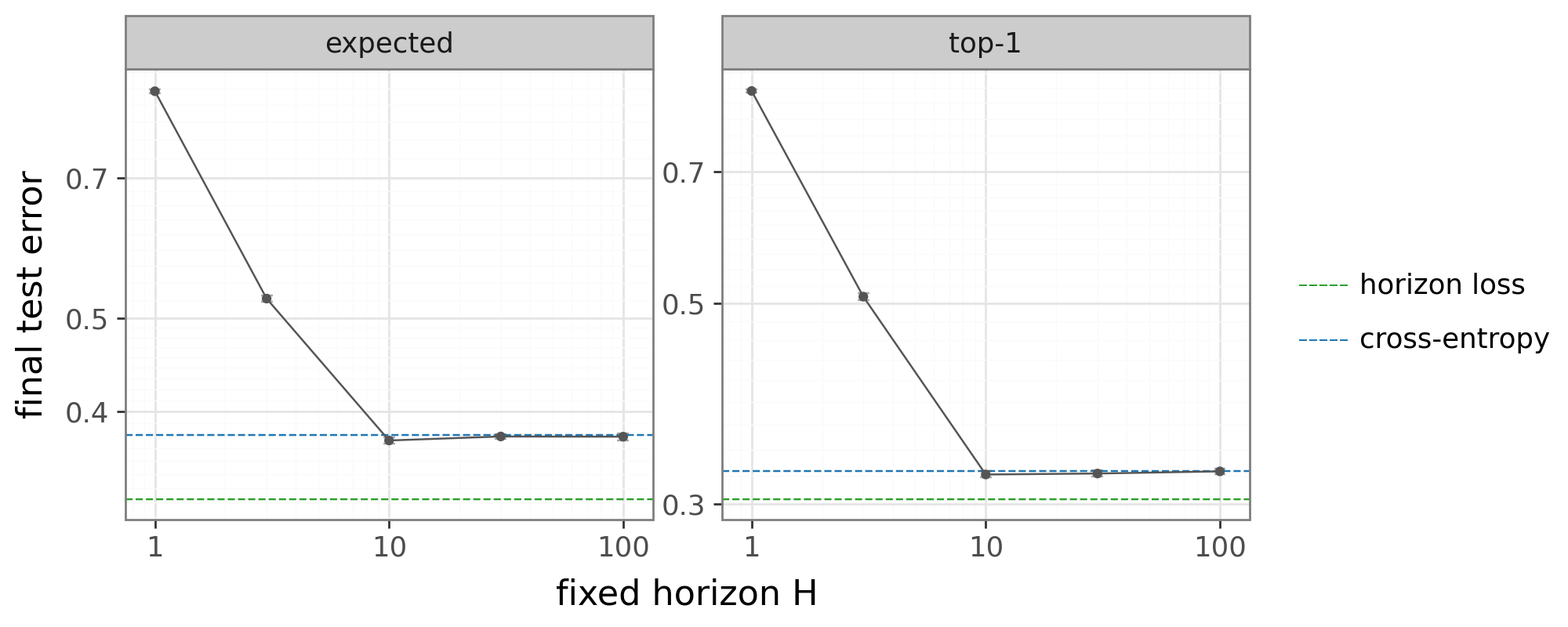}\caption{ResNet-50}\end{subfigure}
\caption{(a) Final test error against $\kappa$ on MNIST for shrinking, constant and growing horizons matched on mean; dashed: cross-entropy. (b) Final test error on ResNet-50 for a fixed horizon $H$; dashed: cross-entropy and the shrinking horizon loss.}
\label{fig:shape}
\end{figure}

If the gain comes from planning to the remaining budget, it should need a shrinking horizon.
A loss that merely upweights some examples, or that plans to a fixed horizon, should not reproduce the gain.
On MNIST we compare the shrinking horizon~\eqref{eq:horizon} with constant and growing horizons matched on their mean, across a range of $\kappa$ (Figure~\ref{fig:shape}a).
Constant and growing horizons give at most a marginal gain over cross-entropy, and lose to it at small $\kappa$.
Only the shrinking horizon gives a consistent gain across the tested range, on expected error at every $\kappa$ and on top-1 error at every $\kappa \ge 1$.

On ResNet-50 we instead hold $H$ fixed throughout training (Figure~\ref{fig:shape}b).
No tested fixed $H$ matches the shrinking horizon.
Fixed horizons $H \ge 10$ tie cross-entropy, since $\sigma(z+H) \approx 1$ for nearly every example.
Horizons $H \le 3$ are myopic from the first step and substantially worse.
The gain comes from a horizon that is long early and short late.

\subsection{Planning Cross-Entropy}
\label{sec:hxent}

\begin{figure}[ht!]
\centering
\begin{subfigure}[t]{0.49\textwidth}\includegraphics[width=\linewidth]{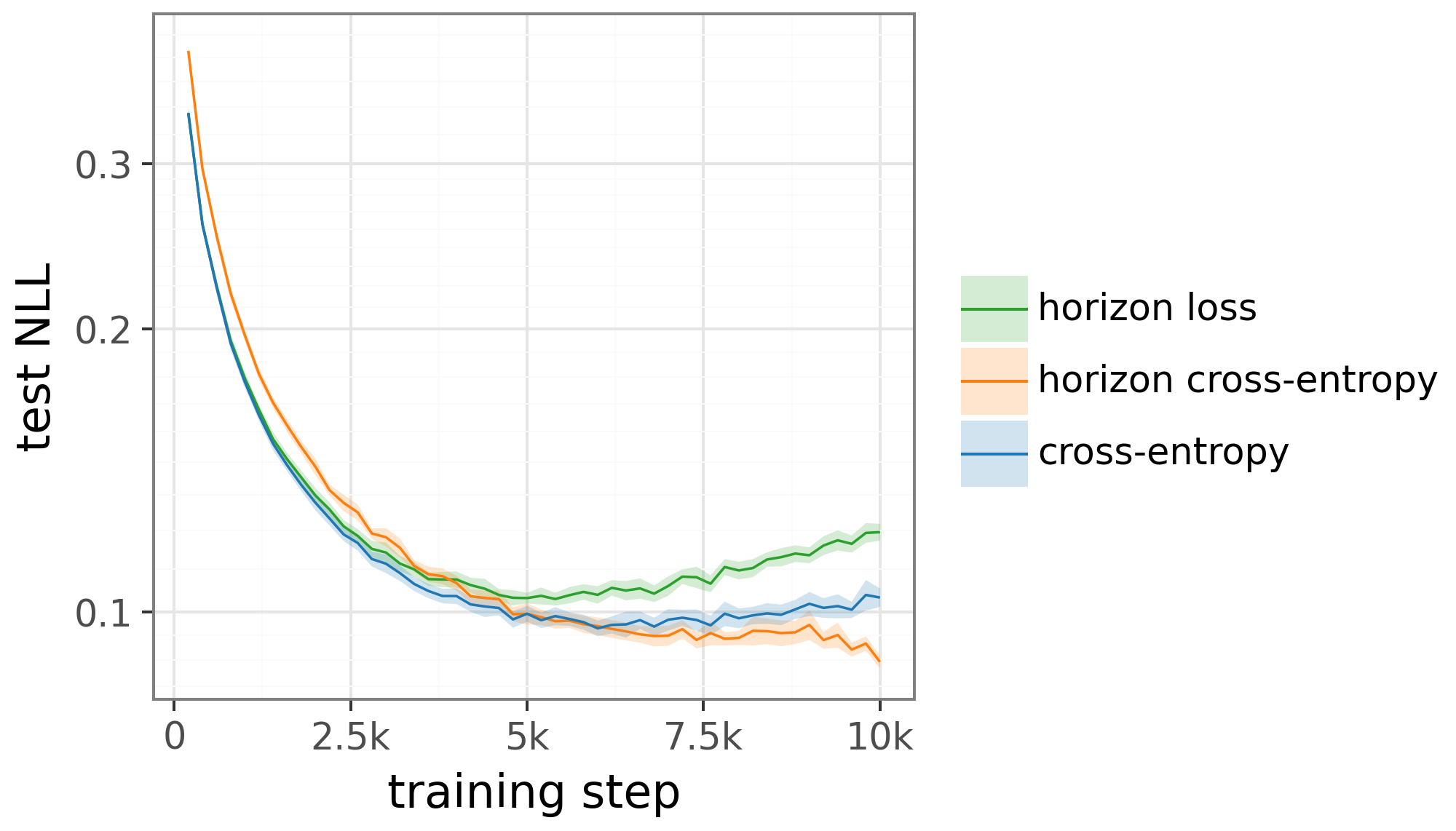}\caption{MNIST (MLP)}\end{subfigure}\hfill
\begin{subfigure}[t]{0.49\textwidth}\includegraphics[width=\linewidth]{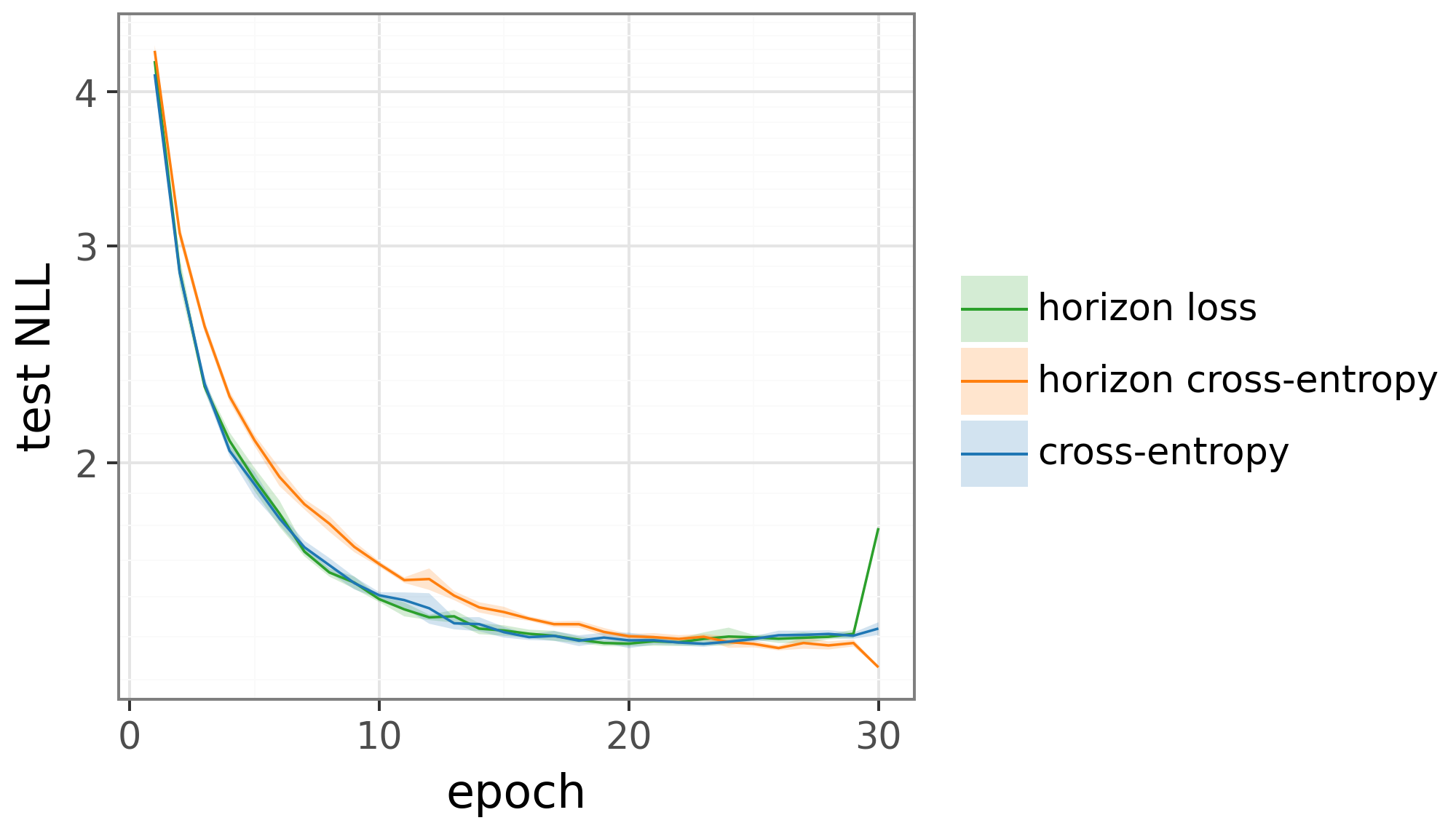}\caption{ImageNet (ResNet-50)}\end{subfigure}
\caption{Test NLL over training at a flat learning rate: horizon cross-entropy ($\kappa = 10$) pulls away from cross-entropy late, as $H_t$ shrinks. The horizon loss, built for accuracy, worsens NLL.}
\label{fig:hxent_nll}
\end{figure}

The planning operator applies to any per-example loss (Section~\ref{sec:horizon}).
Cross-entropy is a proper scoring rule~\citep{gneiting2007strictly}, so its population minimizer is the true label distribution, and on test NLL it is the natural loss to follow as it stands.
Planned the same way, horizon cross-entropy $P_H[\CE]/(1-e^{-H})$, with a capped weight (Appendix~\ref{app:hxent}), lowers test NLL at a flat learning rate, from 0.103 to 0.089 on MNIST and from 1.467 to 1.364 on ResNet-50 (Figure~\ref{fig:hxent_nll}).
A proper loss need not be the best loss to follow on a finite budget.

Horizon cross-entropy moves in the opposite direction to the horizon loss.
NLL is unbounded, so an example far below even odds has the most NLL left to remove, and planning NLL over a long horizon upweights hard examples relative to cross-entropy.
As $H_t \to 0$ the planned loss anneals back to cross-entropy.
Bounded accuracy lets lost causes go, whereas unbounded NLL pays most for them (Appendix~\ref{app:hxent}).

The gains are objective-specific.
Horizon cross-entropy raises ImageNet expected error from 0.378 to 0.393, and the horizon loss raises test NLL, to 0.122 on MNIST and 1.77 on ResNet-50.
Under cosine decay the ImageNet NLL gain shrinks, with 1.270 against 1.289 for cross-entropy.

\section{Related Work}
\label{sec:related}

The tension between a reward objective and a likelihood surrogate predates language models~\citep{ranzato2016sequence,norouzi2016reward,levine2018reinforcement}.
Surrogate-loss theory and proper scoring rules ask which loss has the right minimizer~\citep{bartlett2006convexity,gneiting2007strictly}, whereas we ask how a finite budget should be spent on the way to it.
Closest to our setting, maximum likelihood RL interpolates between expected reward and log-likelihood, moving toward likelihood as more rollouts are sampled per prompt~\citep{tajwar2026maximum}.
The horizon loss makes a similar interpolation, indexed by the remaining budget, so it moves toward expected reward as training runs out.

The same allocation question arises in RL post-training.
For a prompt $q$ with binary reward, $\nabla_\theta p_\theta(q) = p_\theta(q)(1-p_\theta(q))\,\nabla_\theta z_\theta(q)$, where $z_\theta(q)$ is the success log-odds, so the policy gradient carries the weighting of Section~\ref{sec:allocation}.
DAPO and Dr.~GRPO change how the gradient is spread across prompts~\citep{yu2025dapo,liu2025understanding}, but neither uses how much training remains.
Delightful Policy Gradient~\citep{osband2026delightfulpolicygradient,osband2026distributedpolicygradient} also moves the policy gradient toward cross-entropy.
In full-information classification with zero baseline and unit temperature, its expected update is proportional to the negative gradient of the horizon loss with fixed $H = \log 2$ (Appendix~\ref{app:family}).

At fixed $H$, the loss itself is not new.
It is, up to scale, a skew divergence~\citep{lee2001effectiveness} and categorical MAE at $H = 0$, and generalized cross-entropy~\citep{zhang2018generalized} is a mixture over horizons (Appendix~\ref{app:family}).
Bounded losses have been studied for robustness to label noise~\citep{ghosh2017robust}, and dynamics-aware loss already moves a robust loss from fitting toward robustness as training proceeds~\citep{li2023dynamics}.
What we add is a reason for the schedule.
$H$ is the learning budget that remains, and the allocation model says how the loss should change as it shrinks.

Focal loss, curricula, hard-example mining and importance sampling reweight or resample examples by current difficulty or a prescribed schedule~\citep{lin2017focal,bengio2009curriculum,shrivastava2016training,katharopoulos2018not}.
Closest in spirit is RHO-LOSS~\citep{mindermann2022prioritized}, which uses a holdout model to prioritize examples that are learnable and not yet learnt.
The horizon loss uses current success probability and remaining budget, without a holdout model.
Appendix~\ref{app:related} gives more detail and covers meta-gradients and softmax policy gradients.

\section{Discussion}
\label{sec:discussion}

Exact policy gradient values an update by what it buys on the next step, and cross-entropy values it as if learning never ended.
Training has a budget that shrinks with every step.
Planning to that budget moves learning away from lost causes late in training and improves top-1 accuracy at a flat learning rate, most of all with label noise.
The deficit of the exact policy gradient remains with the usual suspects removed, so its allocation problem is not specific to reinforcement learning.

Our analysis strips learning down to allocation.
Examples move independently along a fixed unit-speed path, the horizon is a single scalar $H$, and the conversion from remaining learning rate to logit progress is a calibration $\kappa$ set on test error, which we do not derive from network dynamics.
Deriving the horizon from optimizer and representation dynamics would remove $\kappa$.
The horizon loss is also separable: it prices each example as if the remaining budget were available to that example alone.
A coupled planner can do better, and identical examples expose a failure that no separable rule avoids (Proposition~\ref{prop:ties}).
These simplifications make the finite-horizon effect visible.

Classification also gives information that reinforcement learning usually does not.
The label is known, so $p = \sigma(z)$ and its exact gradient can be computed without sampling actions.
In sampled-action RL, the success probability and the value of further learning must be estimated from partial feedback.
The open question is how to plan to the remaining budget once exploration, credit assignment and action-sampling noise return.

Exploration is a familiar allocation problem in reinforcement learning, giving up reward now for information that pays later.
The horizon loss addresses a less familiar one in ordinary supervised training, where each update trades accuracy now against accuracy later.
That tradeoff is one of many, from which examples deserve a backward pass~\citep{osband2026doesgradientsparkjoy} to how compute and data should be divided across a run.
In each, the learner spends a finite budget over a sequence of updates, so training itself can be posed as a reinforcement-learning problem.

\bibliographystyle{plainnat}

\clearpage
\appendix

\section{Proofs}
\label{app:proofs}

All results use the allocation model of Section~\ref{sec:allocation}.
Table~\ref{tab:notation} collects the notation; the proofs then proceed through the loss family, the two endpoint traps, and the finite-horizon allocation bounds.

\subsection{Notation}
\label{app:notation}

Two clocks appear throughout.
In the allocation model, $\tau$ is budget already spent by a toy flow and $h_\tau = H - \tau$ is budget remaining.
In training, $t$ is the step and $H_t$ estimates the remaining horizon from the learning rate still to come.
Table~\ref{tab:notation} lists every symbol.

\begin{table}[h!]
\centering
\caption{Notation. Toy-flow time $\tau$ and training step $t$ are kept distinct.}
\label{tab:notation}
\begin{tabular}{ll}
\toprule
Symbol & Meaning \\
\midrule
$K$ & number of classes \\
$N$ & number of examples in the allocation model \\
$M$ & number of lost causes in the lost-cause trap \\
$z_i$ & logit gap of example $i$: correct logit minus logsumexp of the rest \\
$p_i = \sigma(z_i)$ & probability of the correct label \\
$H$ & total budget of logit progress in a toy problem \\
$\tau \in [0, H]$ & budget already spent by a toy flow \\
$h_\tau = H - \tau$ & budget remaining in a toy flow \\
$t$, $T$ & training step, total steps \\
$H_t = \kappa \sum_{t \le t' < T} \eta_{t'}$ & estimated remaining horizon in training \\
$V(z) = \sum_i \sigma(z_i)$ & value of a state \\
$V_H^\star(z)$ & optimal final value with budget $H$ \\
$\Delta_H^\star(z) = V_H^\star(z) - V(z)$ & optimal gain \\
$w_A(z, h)$ & weight of rule $A \in \{\mathrm{PG}, \CE, \mathrm{hor}\}$ \\
$J_A$, $\Delta_A = J_A - V(z)$ & final total expected accuracy and gain of rule $A$ \\
$\Delta_A / \Delta_H^\star$ & gain fraction \\
$V_H^\star - J_A$ & regret \\
$C_H(z) = \int_0^H [1 - \sigma(z+s)]\,ds$ & future error over horizon $H$ \\
$L_H(z) = C_H(z)/(1-e^{-H})$ & horizon loss \\
$P_H[\ell](z) = \int_0^H \ell(z+s)\,ds$ & planning operator \\
\bottomrule
\end{tabular}
\end{table}

\subsection{The Loss Family}
\label{app:family}

Write $\rho = e^{-H}$ and $p = \sigma(z)$.
From~\eqref{eq:future},
\[
    C_H(z) = -\log\bigl(\rho + (1-\rho)p\bigr),
\]
so varying $H$ traces a one-parameter family between the zero- and infinite-horizon endpoints.
Equivalently $C_H = P_H[1-\sigma]$; applying the same operator to cross-entropy gives the horizon cross-entropy of Appendix~\ref{app:hxent}.

The endpoints of this family are exact policy gradient and cross-entropy.

\begin{proposition}[Patient-accuracy endpoints]
\label{prop:endpoints}
For $H > 0$ let $C_H(z) = \int_0^H [1-\sigma(z+s)]\,ds$ and $L_H(z) = C_H(z)/(1-e^{-H})$.
Then $C_H(z) = \CE(z) - \CE(z+H)$ and
\[
    -\partial_z L_H(z) = \frac{\sigma(z+H) - \sigma(z)}{1-e^{-H}} = \sigma(z+H)\,(1-\sigma(z)).
\]
Moreover $\lim_{H \to \infty} L_H(z) = \CE(z)$ and $\lim_{H \to 0} L_H(z) = 1-\sigma(z)$, whose negative gradient is $\sigma'(z)$, the policy-gradient weight.
\end{proposition}

\begin{proof}
Since $\tfrac{d}{ds}\CE(z+s) = -(1-\sigma(z+s))$, integrating over $[0, H]$ gives $C_H(z) = \CE(z) - \CE(z+H)$, so $-\partial_z C_H(z) = (1-\sigma(z)) - (1-\sigma(z+H)) = \sigma(z+H) - \sigma(z)$.
Direct computation gives the secant identity
\[
    \sigma(z+H) - \sigma(z)
    = \frac{e^z(e^H - 1)}{(1+e^{z+H})(1+e^z)}
    = \sigma(z+H)\,(1-\sigma(z))\,(1-e^{-H}),
\]
which is the stated gradient after dividing by $1-e^{-H}$.
As $H \to \infty$, $\CE(z+H) \to 0$ and $1-e^{-H} \to 1$, so $L_H \to \CE$.
As $H \to 0$, $C_H(z) = H(1-\sigma(z)) + O(H^2)$ and $1-e^{-H} = H + O(H^2)$, so $L_H \to 1-\sigma(z)$ and $-\partial_z L_H \to \sigma(z)(1-\sigma(z)) = \sigma'(z)$.
\end{proof}

Several fixed members of this family are already known.
At fixed $H$, $L_H$ is a scaled skew divergence~\citep{lee2001effectiveness}.
Generalized cross-entropy~\citep{zhang2018generalized} is a mixture over horizons: for $0 < q < 1$, $(1-p^q)/q = \E_\rho[L_H]$ with $\rho = e^{-H} \sim \mathrm{Beta}(q, 1-q)$.
Here $-\partial_p L_H = 1/(p + (1-p)\rho)$, and the Beta identity $\E[(1+t\rho)^{-1}] = (1+t)^{-q}$ with $t = (1-p)/p$ gives $\E_\rho[1/(p+(1-p)\rho)] = p^{q-1} = -\partial_p (1-p^q)/q$.
Both sides vanish at $p = 1$, so they agree.
The endpoints $q \to 0$ and $q \to 1$ are limits.
Delightful Policy Gradient~\citep{osband2026delightfulpolicygradient} gives another fixed member: its binary expected weight
\[
    \frac{p}{1+p} = \tfrac12\,\sigma(z+\log 2)
\]
is half the horizon factor at $H = \log 2$.
These identities concern fixed losses; the main text sets the horizon from the learning budget that remains.

Finite horizons preserve the Bayes decision and sharpen the probability.

\begin{proposition}[Classification calibration]
\label{prop:calibration}
Let $\mu \in \Delta_K$ be the label posterior at an input, $\rho = e^{-H}$ and $b = \rho/(1-\rho)$.
The population minimizer of $L_H$ over $\pi \in \Delta_K$ is $\pi^\star_k = (c\mu_k - b)_+$, with $c \ge 1$ set by $\sum_k \pi^\star_k = 1$.
It preserves the order of $\mu$ weakly, since classes with small $\mu_k$ can tie at zero, so it makes the Bayes decision, and on its support it widens every gap: $\pi^\star_k - \pi^\star_j = c(\mu_k - \mu_j)$.
For $K = 2$ and an interior optimum, $p^\star = ((1+\rho)\mu - \rho)/(1-\rho)$.
As $H \to \infty$, $\pi^\star \to \mu$; as $H \to 0$, $\pi^\star$ concentrates on the argmax of $\mu$ when it is unique.
\end{proposition}

\begin{proof}
For the correct label, $L_H = -\log(\rho + (1-\rho)\pi_y)/(1-\rho)$, and the constant $1-\rho$ does not affect the minimizer.
The population risk $-\sum_k \mu_k \log(\rho + (1-\rho)\pi_k)$ is convex on $\Delta_K$.
The KKT conditions give $\mu_k(1-\rho)/(\rho + (1-\rho)\pi_k) = \lambda$ wherever $\pi_k > 0$, and $\le \lambda$ wherever $\pi_k = 0$, so $\pi^\star_k = (c\mu_k - b)_+$ with $c = 1/\lambda$.
On the support $S$, $c\,\mu(S) = 1 + |S|\,b$, so $c \ge 1$.
For $K = 2$ with both coordinates positive, $c = 1 + 2b = (1+\rho)/(1-\rho)$, which gives $p^\star$ and $p^\star - \tfrac12 = c(\mu - \tfrac12)$.
As $H \to \infty$, $b \to 0$ and $c \to 1$; as $H \to 0$, $b \to \infty$ and the support shrinks to the argmax.
\end{proof}

This sharpening is why a finite horizon can raise expected accuracy without changing the Bayes classifier (Section~\ref{sec:both}).
The optimum is over the closed simplex, so a softmax reaches its zero entries only in the limit of its logits.
At $H = \infty$ the minimizer is $\mu$, with expected accuracy $\sum_k \mu_k^2 \le \max_k \mu_k$: cross-entropy need not maximize expected accuracy even at its population optimum, yet following its gradient reaches higher expected accuracy within a finite budget.

\subsection{Greed}
\label{app:greed}

We first connect exact policy gradient to the allocation flow, then solve the two examples used in Section~\ref{sec:traps}.

\greedy*

\begin{proof}
$\nabla_\theta V = \sum_i \sigma'(z_i)\nabla_\theta z_i$, so orthonormality gives $\dot z_i = \langle \nabla_\theta z_i, \nabla_\theta V \rangle = \sigma'(z_i)$ along the exact-PG direction.
Rescaling the common step size gives the stated flow.
\end{proof}

Spending each increment of budget on the example with the largest slope $\sigma'(z_i)$ ends in the same place on the polishing trap: from $(a, -2a)$ it spends $a$ on the easy example until $|z_1| = |z_2|$, then the remaining $a$ on the hard one, finishing at $(2a, -a)$.

\reversal*

\begin{proof}
$\sigma$ is increasing, so an optimum spends the whole budget; with $u$ on the easy example the final value is $F(u) = \sigma(a+u) + \sigma(H-2a-u)$ on $[0, H]$.
Since $\sigma'$ is even and strictly decreasing in $|x|$, $F'(u) = \sigma'(a+u) - \sigma'(H-2a-u)$ has the sign of $|H-2a-u| - (a+u)$.
If $H = a$, $F(u) = \sigma(a+u) + \sigma(-(a+u)) = 1$ for every $u$, so every split is optimal.
If $H < a$, then $H-2a-u < 0$ and $|H-2a-u| - (a+u) = a - H > 0$, so $F$ increases and $u = H$ is optimal.
If $a < H \le 3a$: where $H-2a-u \le 0$ the sign is that of $a - H < 0$; where $H-2a-u > 0$ (possible only when $H > 2a$) the sign is that of $(H-3a) - 2u \le 0$.
So $F$ is non-increasing, strictly on a set of full measure, and $u = 0$ is optimal.
Beyond the statement, if $H > 3a$ the same two cases show $F' > 0$ for $u < (H-3a)/2$ and $F' < 0$ after, so the optimum $u = (H-3a)/2$ equalizes the final gaps.
\end{proof}

The polishing trap is the case $H = 2a$, where Proposition~\ref{prop:reversal} puts the whole optimal budget on the hard example.

\begin{proof}[Proof of Proposition~\ref{prop:traps}(i), polishing trap]
Since $1/\sigma'(z) = 2 + 2\cosh z$, the function $G(z) = 2z + 2\sinh z$ satisfies
\[
    \frac{d}{d\tau} G(z_i) = \frac{\dot z_i}{\sigma'(z_i)} = \frac{1}{\sum_j \sigma'(z_j)}.
\]
The right-hand side is common to both examples, so the exact-PG flow conserves $G(z_1) - G(z_2)$.
It also conserves $z_1 + z_2 - \tau$, so at $\tau = 2a$, $z_1 + z_2 = a$.
$G$ is odd, so $(2a, -a)$ satisfies $G(2a) - G(-a) = G(a) - G(-2a)$, and it is the unique solution because $z_1 \mapsto G(z_1) - G(a - z_1)$ is strictly increasing.
With $\sigma(-x) = 1 - \sigma(x)$, $\Delta_{\mathrm{PG}} = 2(\sigma(2a) - \sigma(a))$.
By Proposition~\ref{prop:reversal} ($a < 2a \le 3a$), $\Delta_H^\star = \sigma(0) - \sigma(-2a) = \sigma(2a) - \tfrac12$.
With $x = e^{-a}$,
\[
    \frac{\Delta_{\mathrm{PG}}}{\Delta_H^\star}
    = \frac{2x(1-x)}{(1+x)(1+x^2)} \cdot \frac{2(1+x^2)}{1-x^2}
    = \frac{4x}{(1+x)^2}
    = 4\sigma'(a).
\]
\end{proof}

For the second trap, we first verify that the oracle ignores the lost causes.

\begin{proof}[Optimum of the lost-cause trap]
Any budget $q$ given to the lost causes leaves their gaps in $[-a, H-a] \subset (-\infty, 0]$, where $\sigma$ is convex, so it is best concentrated on one of them.
The remaining value is $g(q) = \sigma(H-q) + \sigma(-a+q)$ plus constants, with $g'(q) = \sigma'(a-q) - \sigma'(H-q) \le 0$ because $a - q \ge H - q \ge 0$.
So $q = 0$ is optimal, and $\Delta_H^\star = \sigma(H) - \sigma(0) = \sigma(H) - \tfrac12$.
\end{proof}

\begin{proof}[Proof of Proposition~\ref{prop:traps}(ii), lost-cause trap]
Along the cross-entropy flow the first gap stays non-negative, so its weight is at most $\tfrac12$.
Each lost gap stays at most $-a + \tau \le H - a \le 0$, so its weight is at least $\tfrac12$.
Hence $\dot z_0 \le \tfrac12 / (\tfrac12 + \tfrac M2) = 1/(M+1)$, and since $\sigma' \le \tfrac14$ the first example gains at most $H/(4M+4)$.
The lost causes receive total progress at most $H$ in a region where $\sigma'(z) \le \sigma(z) \le e^{H-a}$, so they gain at most $He^{H-a}$.
Thus $\Delta_{\CE} \le H/(4M+4) + He^{H-a}$, while $\Delta_H^\star = \sigma(H) - \tfrac12$ is fixed for fixed $H$; the ratio tends to $0$ as $M \to \infty$ and $a - H \to \infty$.
\end{proof}

\subsection{Planning}
\label{app:planning}

The horizon rule is separable: it prices every example as if the remaining budget could be spent on it alone.
The following lemma bounds the cost of that approximation: an example misses its individually planned endpoint only to the extent that horizon weight leaks onto other examples.

\begin{lemma}[Leakage]
\label{lem:leakage}
Let $z(\tau)$ follow the horizon flow on $[0, H]$ with weights $w_j(\tau) = w_{\mathrm{hor}}(z_j(\tau), h_\tau)$.
For any example $i$,
\[
    \sigma(z_i(0) + H) - \sigma(z_i(H)) \le \int_0^H \sum_{j \ne i} w_j(\tau)\,d\tau.
\]
\end{lemma}

\begin{proof}
Write $y_i = z_i + h_\tau$, $S_i = \sum_{j \ne i} w_j$ and $W = \sum_j w_j$.
Then $\dot y_i = w_i/W - 1 = -S_i/W$, and $y_i(0) = z_i(0) + H$, $y_i(H) = z_i(H)$, so
\[
    \sigma(z_i(0) + H) - \sigma(z_i(H)) = \int_0^H \sigma'(y_i)\,\frac{S_i}{W}\,d\tau.
\]
For the horizon rule $w_i = \sigma(y_i)(1-\sigma(z_i))$.
Since $W \ge w_i$ and $y_i \ge z_i$,
\[
    \frac{\sigma'(y_i)}{W} \le \frac{\sigma(y_i)(1-\sigma(y_i))}{\sigma(y_i)(1-\sigma(z_i))} = \frac{1-\sigma(y_i)}{1-\sigma(z_i)} \le 1.
\]
The integrand is therefore at most $S_i$.
\end{proof}

The lemma reduces both regret bounds to controlling distracting weight.
On the polishing trap the distraction is the already-solved example; on the lost-cause trap it is the collection of unreachable examples.

\begin{proof}[Proof of Proposition~\ref{prop:horizon}]
(i) Gaps only increase, so $z_1(H) \ge a$ and $V_H^\star - J_{\mathrm{hor}} = [\sigma(a) - \sigma(z_1(H))] + [\sigma(0) - \sigma(z_2(H))] \le \sigma(z_2(0) + H) - \sigma(z_2(H))$.
The easy example's weight is at most $1 - \sigma(z_1) \le 1 - \sigma(a) \le e^{-a}$, so Lemma~\ref{lem:leakage} bounds the regret by $2ae^{-a}$.
(ii) Lost causes only gain, so $V_H^\star - J_{\mathrm{hor}} \le \sigma(H) - \sigma(z_0(H)) = \sigma(z_0(0) + H) - \sigma(z_0(H))$.
Each lost cause has $z_j(\tau) + h_\tau \le -a + \tau + (H - \tau) = H - a$, so $w_j \le \sigma(H-a) \le e^{H-a}$.
Lemma~\ref{lem:leakage} bounds the regret by $HMe^{H-a}$, which is at most $\epsilon$ once $a \ge H + \log(HM/\epsilon)$.
\end{proof}

Near-optimality on these two traps does not make a separable loss optimal in general.
The next example isolates the obstruction.

\begin{proposition}[Limit of separable allocation]
\label{prop:ties}
Any deterministic separable allocation rule with positive weight $w_i = w(z_i, h) > 0$, locally Lipschitz in $z$ so that its flow is unique, preserves equality between identical examples.
For two examples starting at $(-a, -a)$ with $0 < H \le a$, every such rule therefore splits the budget equally; concentrating it on one example gives strictly larger value, and as $a \to \infty$ with $H$ fixed, equal splitting keeps a fraction $2/(1+e^{H/2})$ of the concentrated gain.
\end{proposition}

\begin{proof}
Swapping two identical examples maps the flow to a flow from the same start, so by uniqueness the flow keeps them identical and each receives $H/2$.
All final gaps lie in $[-a, H-a] \subset (-\infty, 0]$, where $\sigma$ is strictly convex, so $\sigma(-a+H) + \sigma(-a) > 2\sigma(-a+H/2)$.
As $a \to \infty$, $\sigma(x) = e^{x}(1 + o(1))$ on these gaps, so the gain fraction tends to $2(e^{H/2} - 1)/(e^H - 1) = 2/(1 + e^{H/2})$.
\end{proof}

Proposition~\ref{prop:ties} holds for every such separable weighting rule, the horizon loss included.
Matching the oracle $V_H^\star$ here requires breaking the symmetry, by comparing examples jointly, by randomizing, or by ordering them.
The horizon loss is a tractable per-example approximation to the coupled problem, and learning the coupled allocation remains open.

\section{Implementation and Experimental Details}
\label{app:details}

Every method shares the same model, optimizer and schedule within a recipe; only the per-example loss changes.

\subsection{Implementation}

The horizon loss changes only the per-example loss.
For logit gap $z$ and horizon $H$,
\[
    L_H(z) = \frac{C_H(z)}{1-e^{-H}},
    \qquad
    C_H(z) = \CE(z) - \CE(z+H).
\]
For numerical stability we compute $\mathrm{softplus}(z) - \mathrm{softplus}(z+H) = C_H(z) - H$, which has the same gradient, and evaluate the denominator as \texttt{-expm1(-H)}.
On ImageNet the network body runs in bfloat16, but the logits and loss are computed in float32.
At $H = 0$ we use the exact policy-gradient limit.

At training step $t$ we set
\[
    H_t = \mathrm{stopgrad}\Bigl(\kappa \sum_{t \le t' < T} \eta_{t'}\Bigr).
\]
With a flat learning rate this is $\kappa\eta(T-t)$; for the ResNets on ImageNet (150k steps at $10^{-3}$), $H_0 \approx 150\kappa$.
The scalar $\kappa$ converts remaining learning rate into logit progress and is calibrated per dataset and schedule.
For $K$ classes the logit gap is~\eqref{eq:gap}.

Horizon cross-entropy uses the same construction.
Its normalized loss $P_H[\CE]/(1-e^{-H})$ has negative gradient $C_H/(1-e^{-H})\,\nabla z$, and we implement it by multiplying the cross-entropy gradient by the detached factor $C_H/((1-e^{-H})(1-p))$, capped at 10.
The trained weight on $\nabla z$ is therefore $\min\{C_H/(1-e^{-H}),\, 10(1-p)\}$, which equals the uncapped weight whenever the cap is inactive.

\begin{algorithm}[h]
\caption{Horizon loss}
\label{alg:horizon}
\begin{algorithmic}[1]
\Require logits $f$, label $y$, step $t$, LR schedule $\eta$, calibration $\kappa$
\State $z \gets f_y - \log \sum_{k \ne y} e^{f_k}$ \Comment{logit gap}
\State $H \gets \mathrm{stopgrad}\big(\kappa \sum_{t \le t' < T} \eta_{t'}\big)$ \Comment{remaining LR sum, in logit units}
\If{$H = 0$} \State \Return $-\sigma(z)$ \Comment{exact policy-gradient limit} \EndIf
\State $D \gets \mathrm{softplus}(z) - \mathrm{softplus}(z+H)$ \Comment{$= C_H - H$; same gradient}
\State \Comment{for $H \le 0.1$: $D = -\mathrm{log1p}\big((e^H - 1)\,\sigma(z)\big)$}
\State \Return $D / (1 - e^{-H})$ \Comment{\texttt{-expm1(-H)} in code}
\end{algorithmic}
\end{algorithm}

\subsection{Selection and Reporting}

Within each training recipe, only the loss changes across methods.
We report final test expected accuracy $\E[p_y]$ and top-1 accuracy, as mean $\pm$ SEM over seeds in tables and mean $\pm 2$ SEM in figures.
For relative error reductions (Figure~\ref{fig:summary}), the SEM of the ratio is propagated from the two independent means.
SEM reflects variation across seeds, not uncertainty from model or hyperparameter selection, and with three seeds $\pm 2$ SEM is only a rough guide.
We do not report significance tests.

\begin{figure}[ht!]
\centering
\begin{subfigure}{0.49\textwidth}\includegraphics[width=\linewidth]{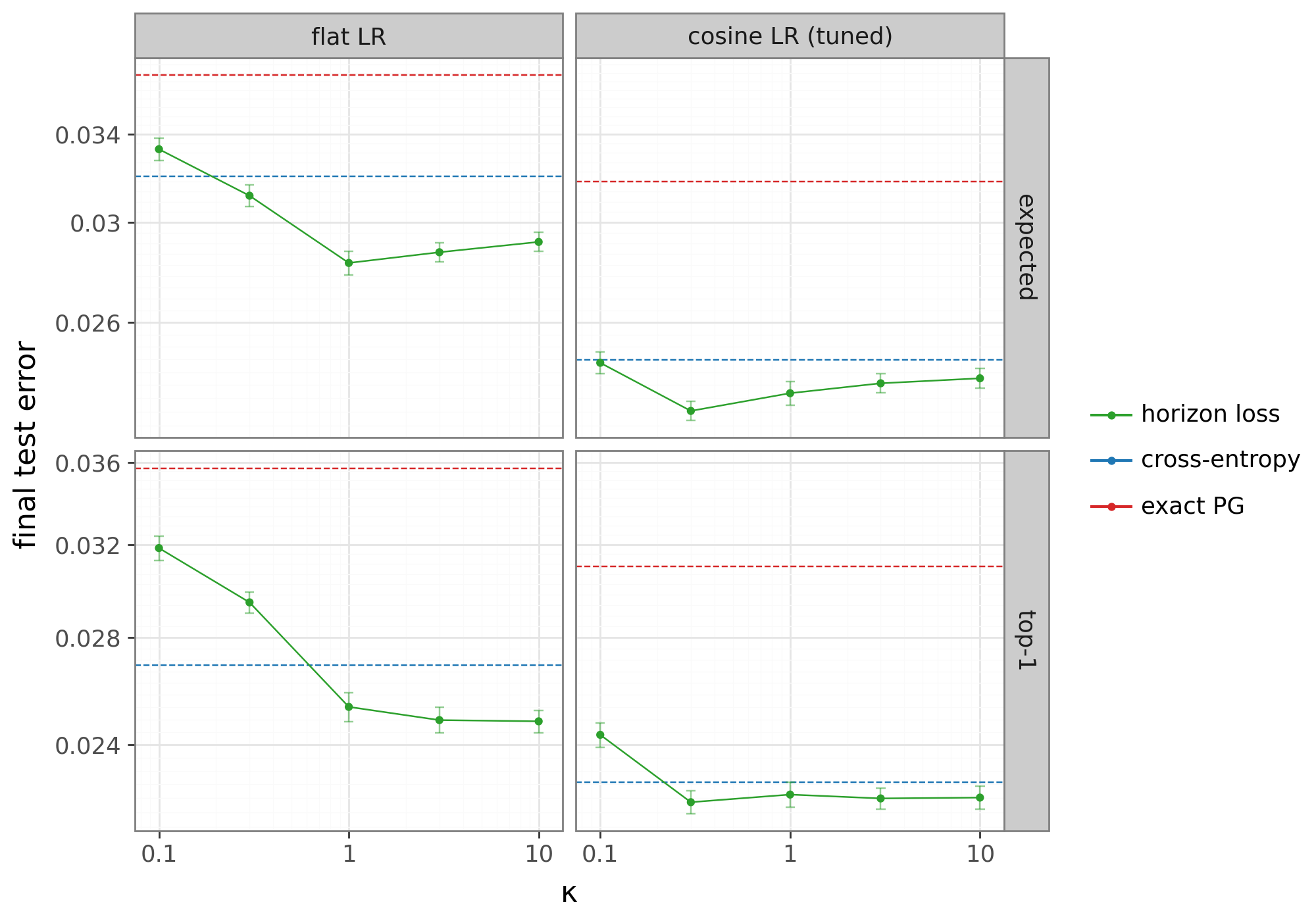}\caption{MNIST}\end{subfigure}\hfill
\begin{subfigure}{0.49\textwidth}\includegraphics[width=\linewidth]{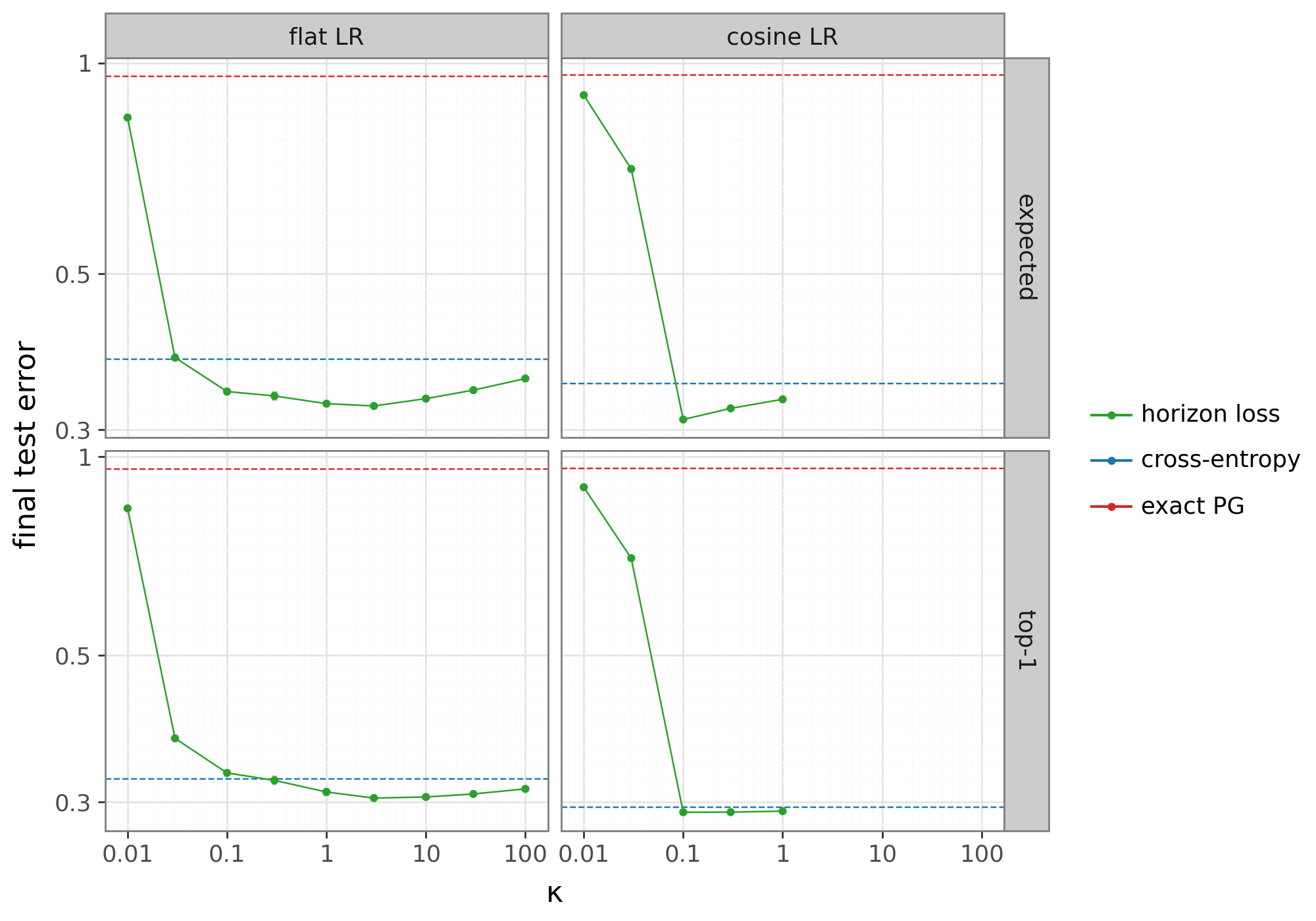}\caption{ResNet-50}\end{subfigure}
\caption{Final expected and top-1 error against $\kappa$, flat and cosine schedules.}
\label{fig:cosine}
\end{figure}

There is no separate validation split, and every selection below uses test metrics.
Expected and top-1 accuracy for a setting always come from the same configuration.
\begin{itemize}
\item \textbf{MNIST, flat learning rate.} Every method uses the shared learning rate $10^{-3}$. The horizon loss uses $\kappa = 1$, reported from the same 30-seed sweep that also contains the $\kappa$ grid $\{0.01, 0.03, \ldots, 100\}$ (half-decade steps) and the switch fractions $\{0.01, \ldots, 0.99\}$. The learning-rate grid $\{1, 3, 10, 30, 100\} \times 10^{-4}$ is a separate robustness check (Figure~\ref{fig:lr}a) and is not used for selection.
\item \textbf{MNIST, cosine decay.} Each method uses its own best peak learning rate from $\{1, 3, 10\} \times 10^{-3}$, chosen on expected accuracy only. The horizon loss uses $\kappa = 0.3$, the ImageNet cosine value, with $\kappa \in \{0.1, 0.3, 1, 3, 10\}$ shown in Figure~\ref{fig:cosine}.
\item \textbf{ImageNet, flat learning rate.} Every method uses the shared learning rate $10^{-3}$ ($5 \times 10^{-4}$ for ViT-S/16), without per-method tuning. We chose $\kappa = 3$ from the grid $\{0.01, 0.03, \ldots, 100\}$ on ResNet-50 over 3 seeds, then reran cross-entropy and $\kappa = 3$ over 10 seeds for the headline. The 10 headline seeds include the 3 selection seed indices. The value is reused unchanged for ResNet-101, ViT-S/16 and every noise level.
\item \textbf{ImageNet, cosine decay.} We chose $\kappa = 0.3$ from $\{0.01, 0.03, 0.1, 0.3, 1\}$ on ResNet-50 over 3 seeds, with the same 10-seed rerun and reuse. On those seeds $\kappa = 0.1$ is better on expected accuracy (69.0 vs 67.8\%) and ties on top-1, but it sits one grid step from the collapse at $\kappa = 0.03$, so we kept the more conservative value.
\item \textbf{Learning-rate sweeps.} On ResNet-50, cross-entropy and exact PG are swept over $\{0.5, 1, 2\} \times 10^{-3}$, and exact PG further over $10^{-4}$--$10^{-2}$. These sweeps assess sensitivity to the shared learning rate and are not used for selection. Switch fractions are $\{0.5, 0.7, 0.8, 0.9, 0.95\}$.
\end{itemize}
For MNIST, $\kappa = 1$ is also reused for the noise, width and training-budget sweeps.
Every tested $\kappa$ from $0.3$ to $100$ improves expected error over cross-entropy on MNIST and ResNet-50 at a flat learning rate (Figure~\ref{fig:kappa}), and $\kappa = 100$ retains more than 40\% of the ImageNet gain.

\subsection{MNIST}

We use an MLP with two hidden layers of width 50 and ReLU activations on raw pixels scaled to $[0,1]$, with no augmentation.
We train with Adam, batch size 100 sampled uniformly with replacement, 100 warmup steps, a flat learning rate of $10^{-3}$, 10k training steps and 30 seeds, with $\kappa = 1$, so $H_0 \approx 10\kappa$.
Cosine decay ends at 1\% of the peak learning rate; under cosine decay we use $\kappa = 0.3$ and each method at its best peak learning rate.
For the label-noise experiments we replace each training label, with probability equal to the noise rate from 0 to 50\%, by a uniformly chosen wrong class; a fixed noise seed gives every method and seed the same corrupted labels, and test labels stay clean.
Figure~\ref{fig:noise}c tracks the probability assigned to the corrupted label on the flipped training examples.

\subsection{ImageNet}

ResNet-50 and ResNet-101 use AdamW with zero weight decay and a flat learning rate of $10^{-3}$ after one warmup epoch.
ViT-S/16 uses a learning rate of $5 \times 10^{-4}$ and five warmup epochs.
All models train for 30 epochs with batch size 256, bfloat16 arithmetic and no label smoothing, over 3--10 seeds.
Training uses random-resized crops (area $0.08$--$1$, aspect ratio $3/4$--$4/3$) and horizontal flips at $224 \times 224$; evaluation uses a center crop.
Cosine decay ends at a learning rate of $10^{-5}$, and the remaining horizon $H_t$ follows the decayed schedule.
Label noise replaces each training label, with probability equal to the noise rate, by a uniformly chosen wrong class, keyed on a hash of the image so that every epoch, seed and method sees the same corrupted label.
We use $\kappa = 3$ for flat schedules and $\kappa = 0.3$ for cosine, selected on ResNet-50 and reused across architectures.
The weight-decay control uses cosine decay with weight decay $0.05$.
Horizon cross-entropy uses its own calibration, $\kappa = 10$.

\section{Robustness, Baselines and Extensions}
\label{app:robustness}

Unless stated otherwise, the protocol matches Section~\ref{sec:experiments}.

\subsection{Full Results}

Tables~\ref{tab:mnist}--\ref{tab:imagenet_cosine} report final expected and top-1 accuracy, with $n$ the number of seeds.

\begin{table}[h]
\centering
\caption{MNIST, flat learning rate: final test accuracy (\%), mean $\pm$ SEM over $n$ seeds. The switch trains with cross-entropy and switches to exact PG for the last 10\% of steps. Bold: best in column.}
\label{tab:mnist}
\begin{tabular}{lccc}
\toprule
Method & Expected & Top-1 & $n$ \\
\midrule
cross-entropy & $96.80 \pm 0.03$ & $97.31 \pm 0.03$ & 30 \\
exact PG & $96.30 \pm 0.31$ & $96.43 \pm 0.31$ & 30 \\
switch at 0.9 & $97.12 \pm 0.02$ & $\mathbf{97.52 \pm 0.02}$ & 30 \\
horizon loss, $\kappa=1$ & $\mathbf{97.17 \pm 0.02}$ & $97.47 \pm 0.03$ & 30 \\
\bottomrule
\end{tabular}
\end{table}

\begin{table}[h]
\centering
\caption{MNIST, cosine decay: final test accuracy (\%), mean $\pm$ SEM over $n$ seeds, each method at its best peak learning rate. Bold: best in column.}
\label{tab:mnist_cosine}
\begin{tabular}{lccc}
\toprule
Method & Expected & Top-1 & $n$ \\
\midrule
cross-entropy & $97.54 \pm 0.02$ & $97.73 \pm 0.02$ & 30 \\
exact PG & $96.82 \pm 0.31$ & $96.90 \pm 0.31$ & 30 \\
horizon loss, $\kappa=0.3$ & $\mathbf{97.71 \pm 0.02}$ & $\mathbf{97.79 \pm 0.02}$ & 30 \\
\bottomrule
\end{tabular}
\end{table}

\begin{table}[h]
\centering
\caption{ImageNet, flat learning rate: final test accuracy (\%), mean $\pm$ SEM over $n$ seeds. The calibrated value is $\kappa = 3$. Bold: best within each model.}
\label{tab:imagenet}
\begin{tabular}{llccc}
\toprule
Model & Method & Expected & Top-1 & $n$ \\
\midrule
ResNet-50 & cross-entropy & $62.1 \pm 0.17$ & $67.4 \pm 0.17$ & 10 \\
 & exact PG & $4.2 \pm 0.31$ & $4.3 \pm 0.31$ & 3 \\
 & switch at 0.9 & $65.8 \pm 0.17$ & $66.7 \pm 0.17$ & 3 \\
 & horizon loss, $\kappa=1$ & $67.3 \pm 0.04$ & $68.9 \pm 0.05$ & 10 \\
 & horizon loss, $\kappa=3$ & $\mathbf{67.5 \pm 0.04}$ & $\mathbf{69.6 \pm 0.05}$ & 10 \\
\midrule
ResNet-101 & cross-entropy & $64.4 \pm 0.37$ & $69.0 \pm 0.30$ & 3 \\
 & horizon loss, $\kappa=1$ & $69.0 \pm 0.04$ & $70.5 \pm 0.05$ & 3 \\
 & horizon loss, $\kappa=3$ & $\mathbf{69.3 \pm 0.03}$ & $\mathbf{71.4 \pm 0.03}$ & 3 \\
\midrule
ViT-S/16 & cross-entropy & $59.4 \pm 0.17$ & $65.0 \pm 0.12$ & 3 \\
 & horizon loss, $\kappa=1$ & $63.2 \pm 0.11$ & $64.7 \pm 0.12$ & 3 \\
 & horizon loss, $\kappa=3$ & $\mathbf{64.1 \pm 0.15}$ & $\mathbf{66.0 \pm 0.19}$ & 3 \\
\bottomrule
\end{tabular}
\end{table}

\begin{table}[h]
\centering
\caption{ImageNet, cosine decay: final test accuracy (\%), mean $\pm$ SEM over $n$ seeds. WD: weight decay $0.05$. The calibrated value is $\kappa = 0.3$. Bold: best within each setting.}
\label{tab:imagenet_cosine}
\begin{tabular}{llccc}
\toprule
Setting & Method & Expected & Top-1 & $n$ \\
\midrule
ResNet-50 & cross-entropy & $65.2 \pm 0.02$ & $70.5 \pm 0.02$ & 10 \\
 & switch at 0.9 & $\mathbf{68.3 \pm 0.06}$ & $\mathbf{71.2 \pm 0.03}$ & 3 \\
 & switch at 0.95 & $67.5 \pm 0.06$ & $71.0 \pm 0.06$ & 3 \\
 & horizon loss, $\kappa=0.3$ & $67.8 \pm 0.04$ & $71.1 \pm 0.05$ & 10 \\
 & horizon loss, $\kappa=1$ & $66.8 \pm 0.02$ & $70.9 \pm 0.02$ & 10 \\
\midrule
ResNet-50 + WD & cross-entropy & $66.6 \pm 0.03$ & $73.3 \pm 0.04$ & 3 \\
 & horizon loss, $\kappa=0.3$ & $\mathbf{69.3 \pm 0.02}$ & $73.5 \pm 0.10$ & 3 \\
 & horizon loss, $\kappa=1$ & $68.2 \pm 0.05$ & $\mathbf{73.5 \pm 0.04}$ & 3 \\
\midrule
ResNet-101 & cross-entropy & $67.7 \pm 0.02$ & $72.2 \pm 0.08$ & 3 \\
 & horizon loss, $\kappa=0.3$ & $\mathbf{69.8 \pm 0.05}$ & $\mathbf{72.8 \pm 0.07}$ & 3 \\
 & horizon loss, $\kappa=1$ & $68.9 \pm 0.02$ & $72.5 \pm 0.05$ & 3 \\
\midrule
ViT-S/16 & cross-entropy & $62.3 \pm 0.09$ & $68.1 \pm 0.15$ & 3 \\
 & horizon loss, $\kappa=0.3$ & $\mathbf{65.7 \pm 0.14}$ & $\mathbf{68.7 \pm 0.12}$ & 3 \\
 & horizon loss, $\kappa=1$ & $64.6 \pm 0.07$ & $68.3 \pm 0.05$ & 3 \\
\bottomrule
\end{tabular}
\end{table}

\subsection{Learning Rate}

\begin{figure}[ht!]
\centering
\begin{subfigure}{0.49\textwidth}\includegraphics[width=\linewidth]{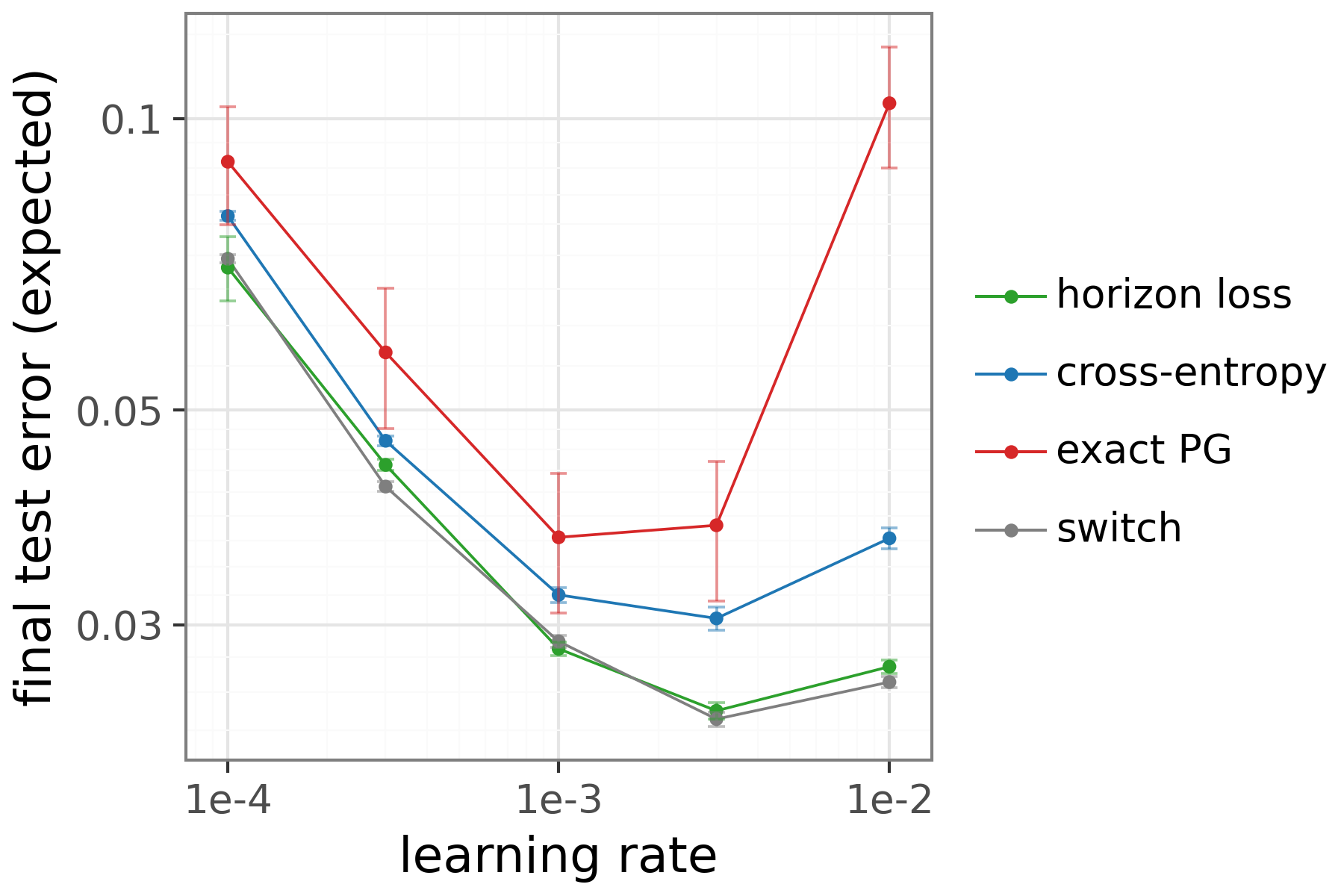}\caption{MNIST}\end{subfigure}\hfill
\begin{subfigure}{0.49\textwidth}\includegraphics[width=\linewidth]{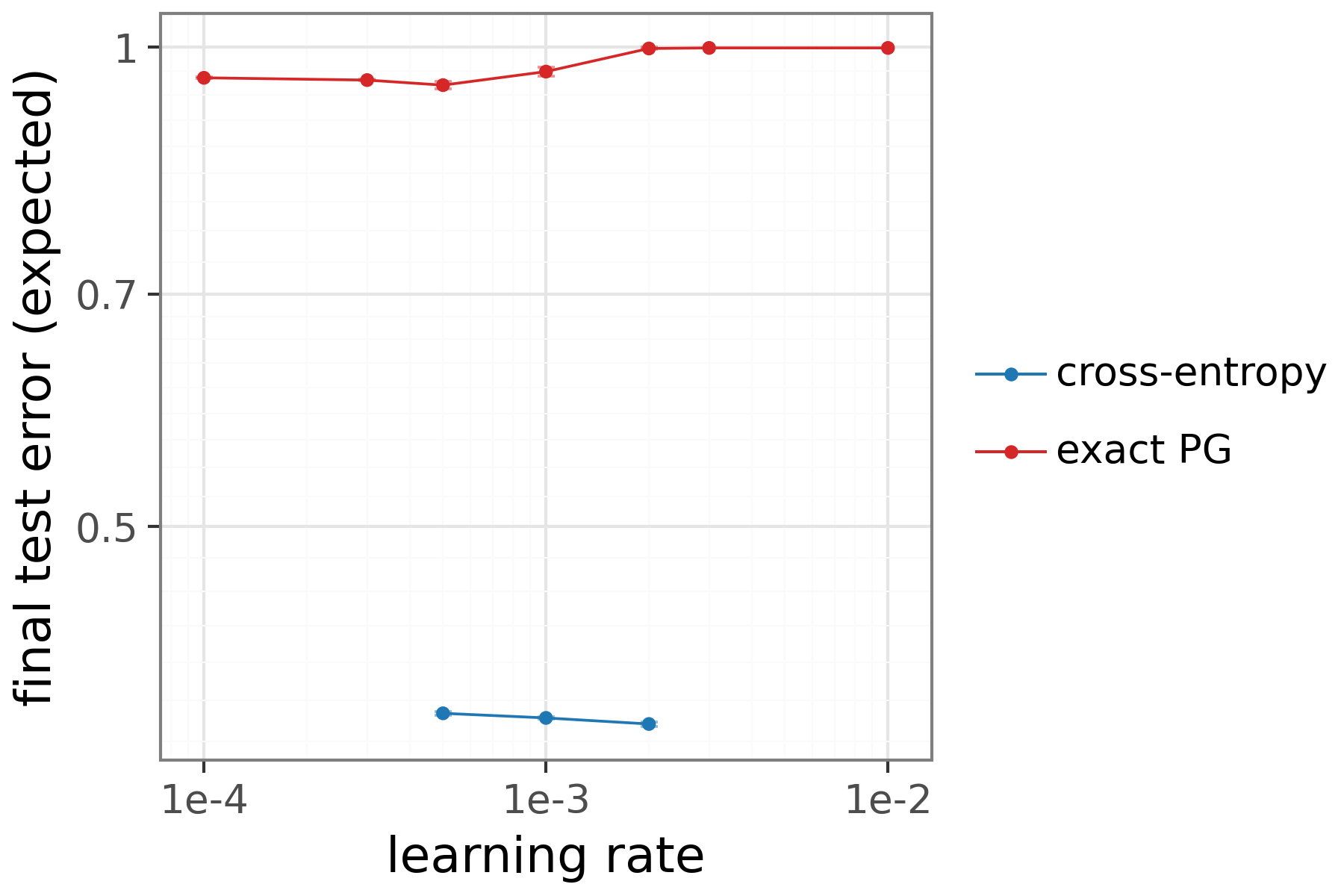}\caption{ResNet-50}\end{subfigure}
\caption{Final error against learning rate. (a) MNIST, every method. (b) ResNet-50, cross-entropy and exact PG.}
\label{fig:lr}
\end{figure}

On MNIST, cross-entropy is best near $3 \times 10^{-3}$ and exact PG near $10^{-3}$.
The horizon loss beats cross-entropy at every tested learning rate (Figure~\ref{fig:lr}a).

On ResNet-50, the cross-entropy sweep covers $5 \times 10^{-4}$ to $2 \times 10^{-3}$ and is best at the edge of the grid, so we do not claim it is fully tuned.
Exact PG stays below 5\% expected accuracy at every learning rate from $10^{-4}$ to $10^{-2}$ (Figure~\ref{fig:lr}b).

\subsection{Horizon Calibration and Switches}

\begin{figure}[ht!]
\centering
\begin{subfigure}{0.49\textwidth}\includegraphics[width=\linewidth]{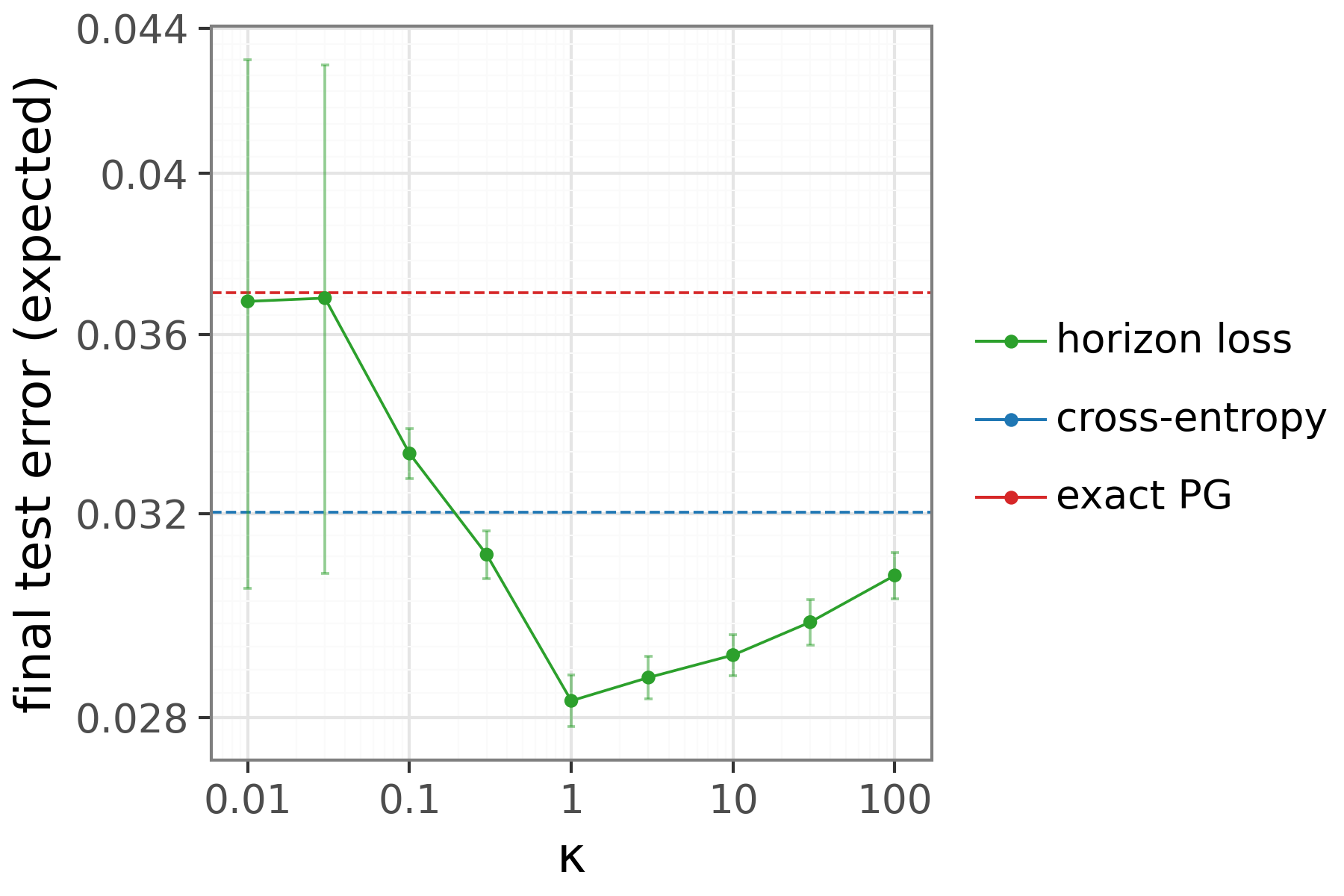}\caption{MNIST}\end{subfigure}\hfill
\begin{subfigure}{0.49\textwidth}\includegraphics[width=\linewidth]{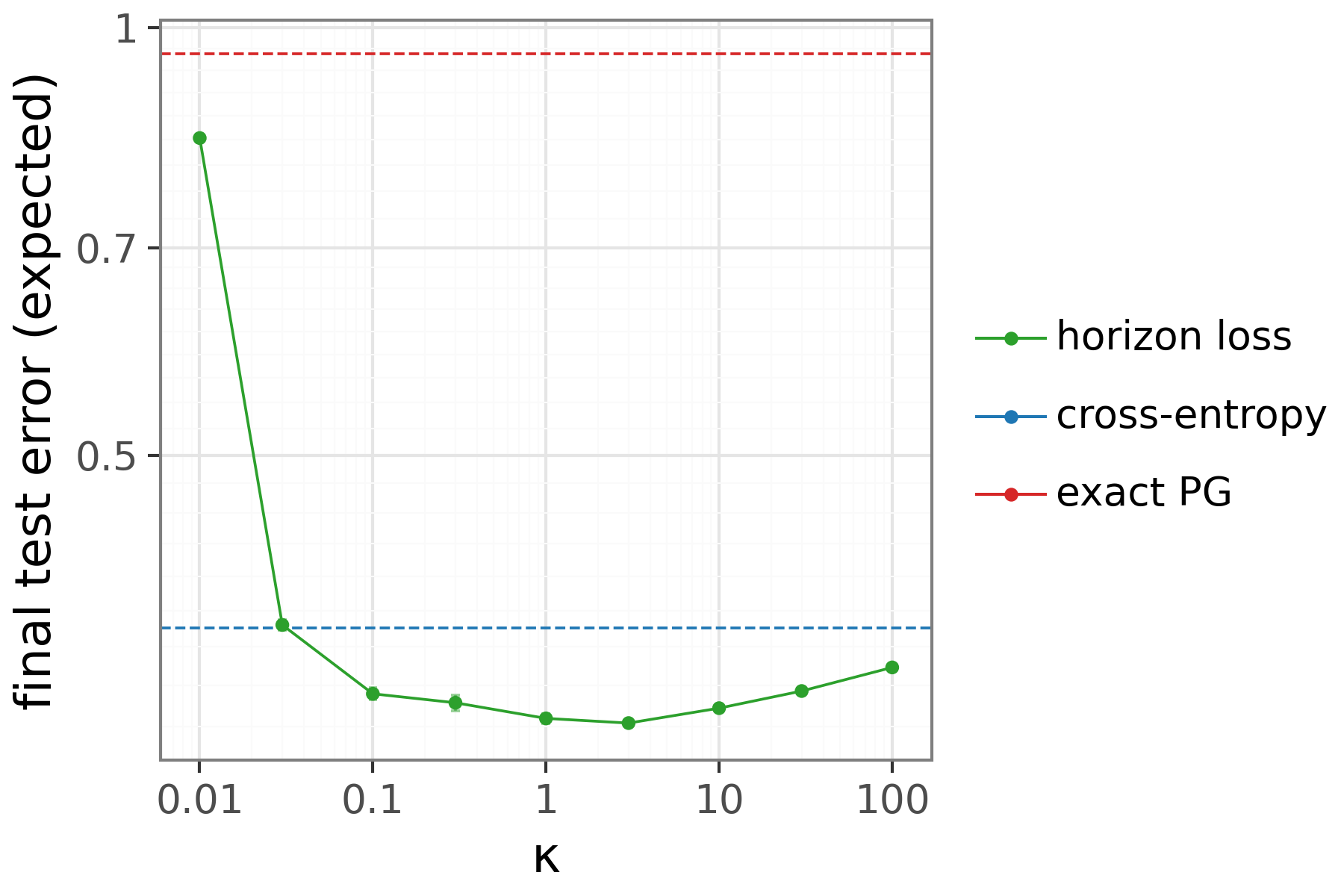}\caption{ResNet-50}\end{subfigure}
\caption{Final error against $\kappa$. Dashed lines: cross-entropy and exact PG.}
\label{fig:kappa}
\end{figure}

The basin in $\kappa$ is broad on both datasets (Figure~\ref{fig:kappa}).
At $\kappa = 1$, away from the calibrated value, the ImageNet top-1 gain over cross-entropy is 1.5 points on both ResNets and $-0.3$ on ViT-S/16 (Table~\ref{tab:imagenet}).

\begin{figure}[ht!]
\centering
\begin{subfigure}{0.49\textwidth}\includegraphics[width=\linewidth]{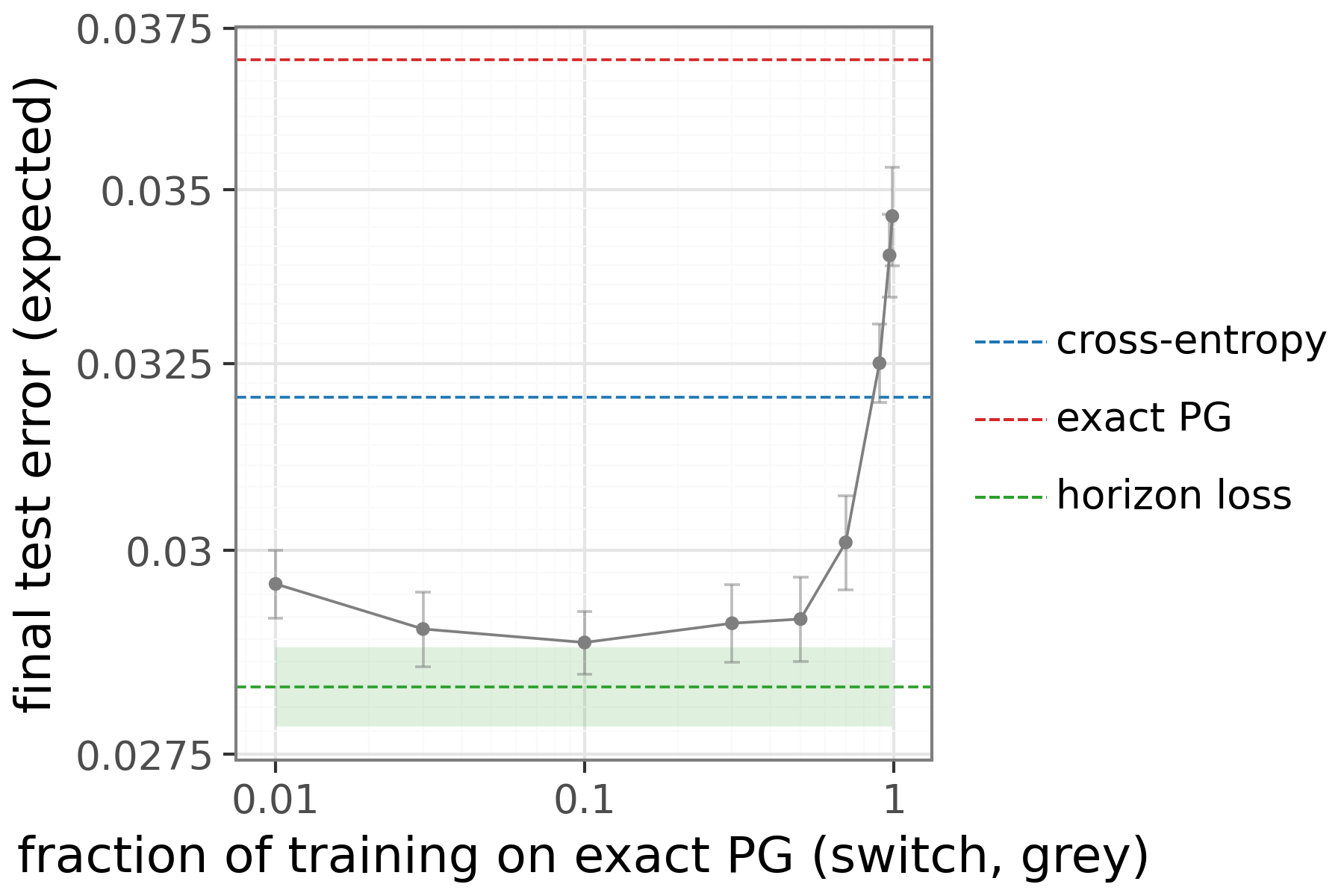}\caption{MNIST}\end{subfigure}\hfill
\begin{subfigure}{0.49\textwidth}\includegraphics[width=\linewidth]{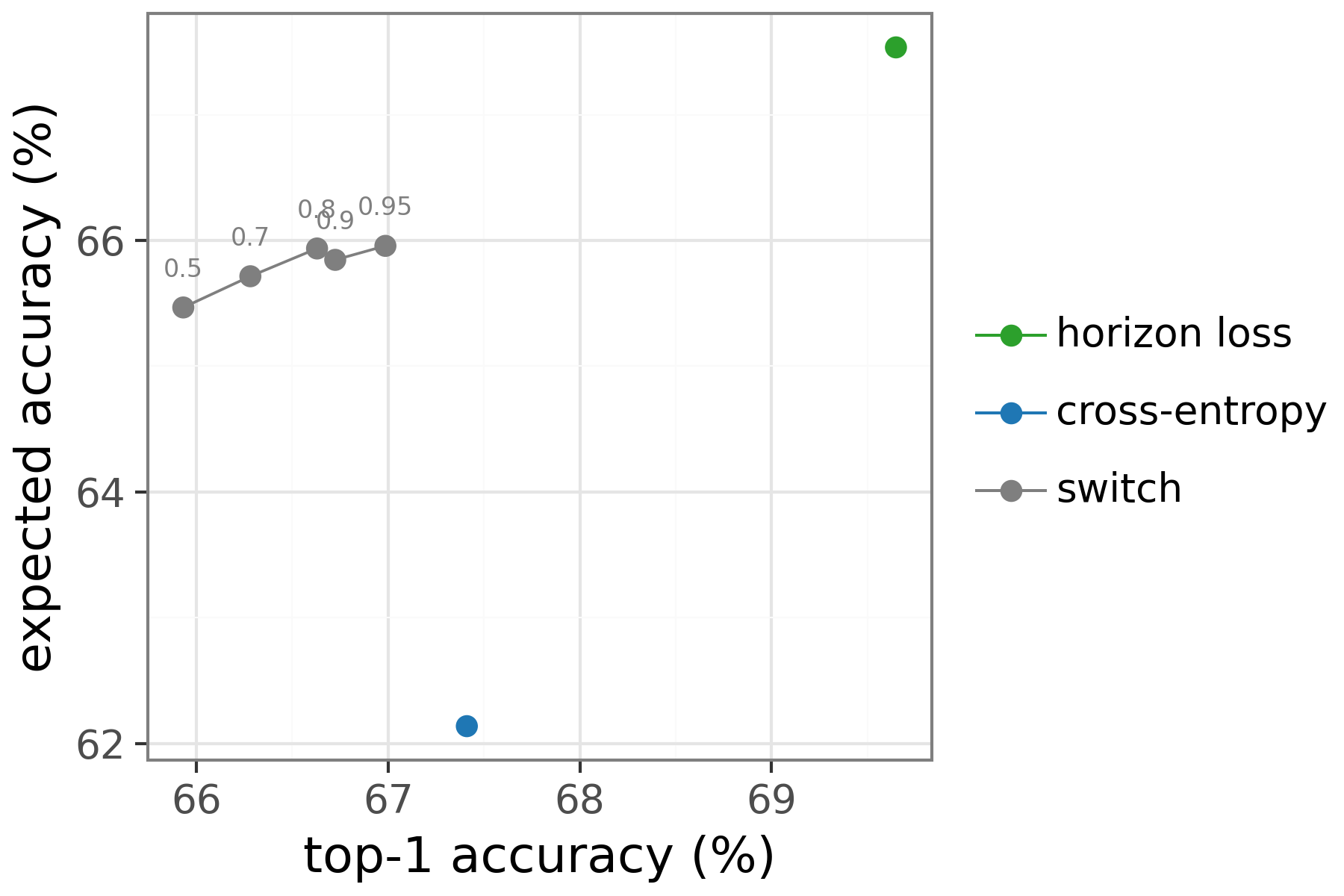}\caption{ResNet-50}\end{subfigure}
\caption{Late switch from cross-entropy to exact PG. (a) MNIST, final error against the fraction of training spent on exact PG, so a switch at 0.9 in Table~\ref{tab:mnist} sits at 0.1. (b) ResNet-50, expected against top-1 accuracy.}
\label{fig:switch}
\end{figure}

A late switch from cross-entropy to exact PG is a hand-built horizon: infinite before the switch, zero after it.
On MNIST, a tuned switch roughly matches the horizon loss (Figure~\ref{fig:switch}a).
On ResNet-50 at a flat learning rate, the switch trades top-1 accuracy for expected accuracy (Figure~\ref{fig:switch}b); under cosine decay, a switch at 0.9 slightly exceeds the horizon loss on both metrics (Table~\ref{tab:imagenet_cosine}).

\subsection{Focal Loss}

Focal loss multiplies cross-entropy by $(1-p)^\gamma$, a static function of the current probability~\citep{lin2017focal}.
On ResNet-50 at a flat learning rate, every $\gamma \in \{1,2,5\}$ has higher expected and top-1 error than cross-entropy.
Horizon cross-entropy beats focal loss on all three metrics, including NLL (1.364 against at least 1.39).

\begin{figure}[ht!]
\centering
\includegraphics[width=0.85\textwidth]{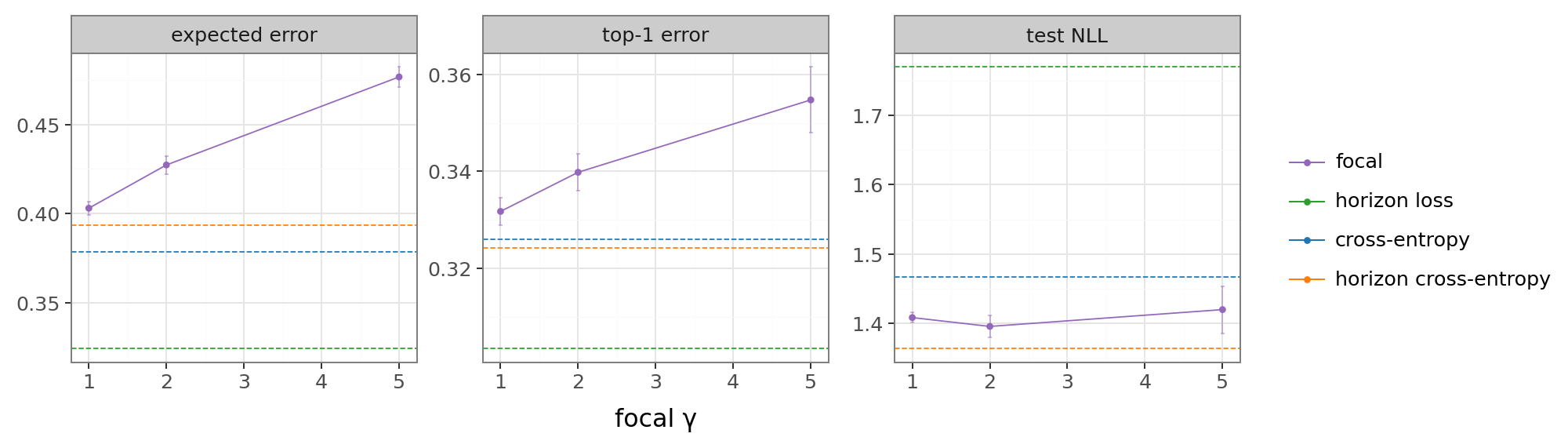}
\caption{Expected error, top-1 error and test NLL against focal $\gamma$, ResNet-50, flat learning rate. Dashed lines: cross-entropy, horizon loss and horizon cross-entropy.}
\label{fig:focal}
\end{figure}

\subsection{Learning-Rate Decay and Weight Decay}

On ImageNet, the top-1 gain over cross-entropy is about 0.6 points on all three models under cosine decay, and 0.2 points with weight decay added.
On MNIST under cosine decay, the horizon loss reduces expected error by 7\% and top-1 error by 3\%; the top-1 interval is marginal.

\subsection{Width and Training Budget}

\begin{figure}[ht!]
\centering
\begin{subfigure}{0.32\textwidth}\includegraphics[width=\linewidth]{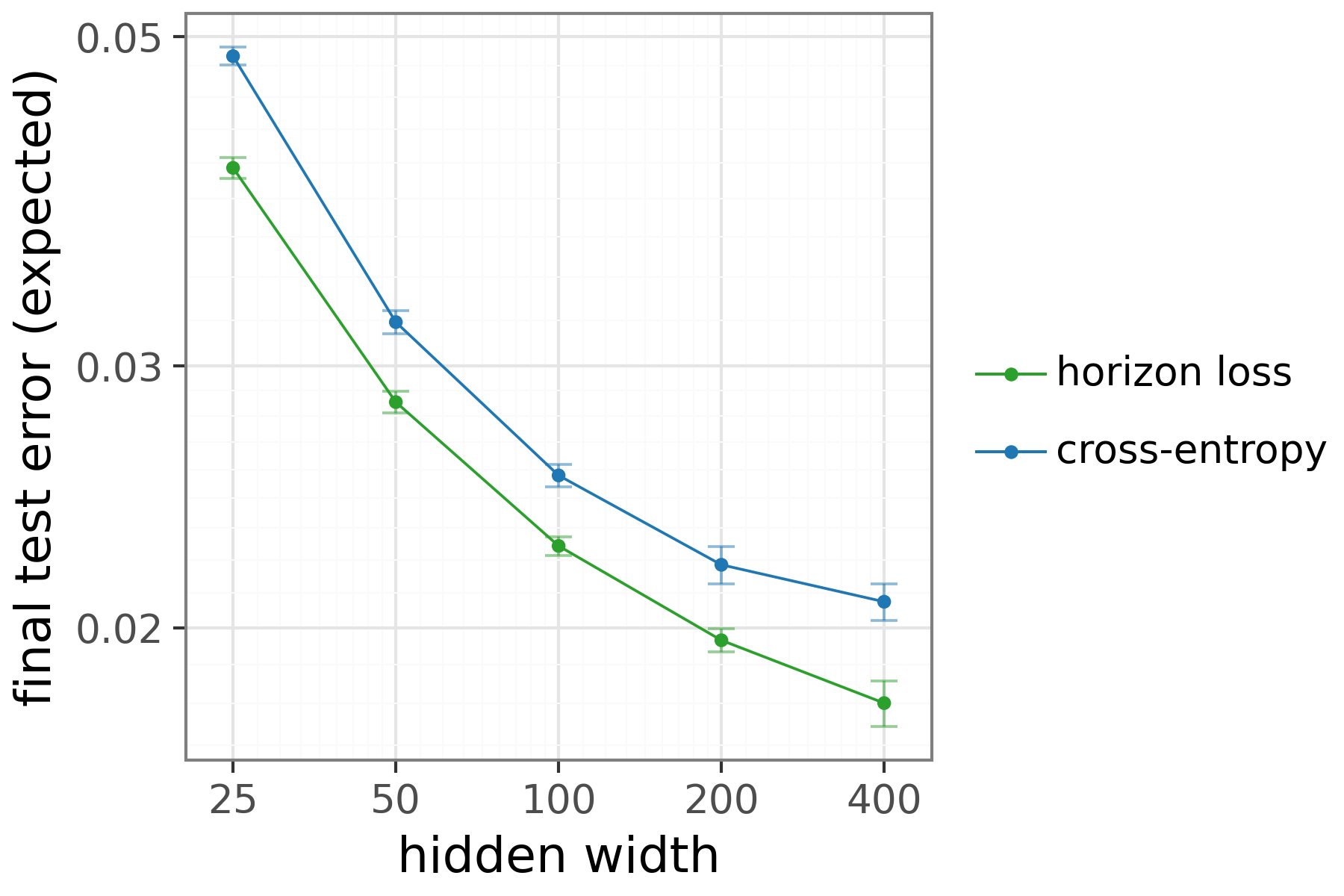}\caption{}\end{subfigure}\hfill
\begin{subfigure}{0.32\textwidth}\includegraphics[width=\linewidth]{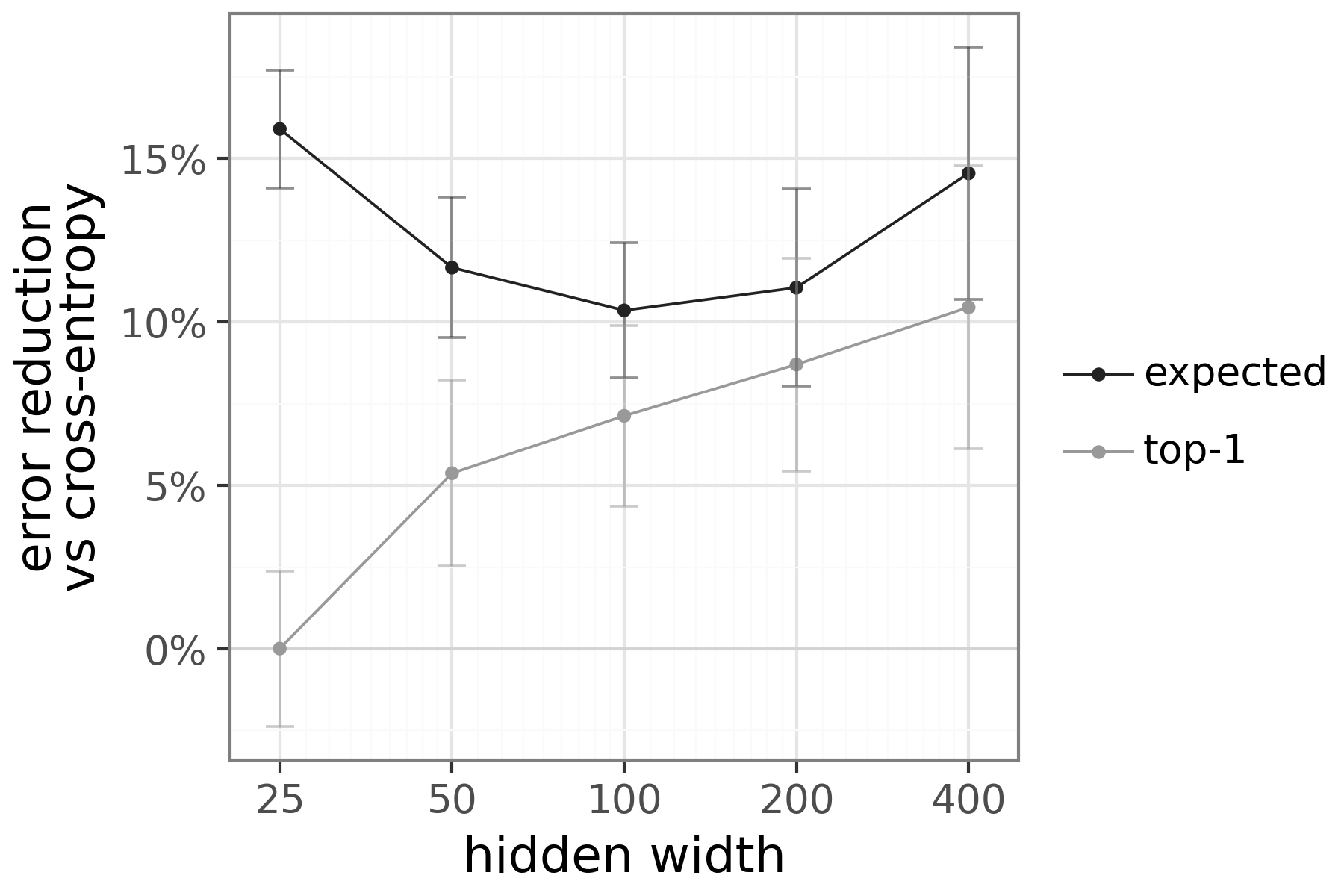}\caption{}\end{subfigure}\hfill
\begin{subfigure}{0.32\textwidth}\includegraphics[width=\linewidth]{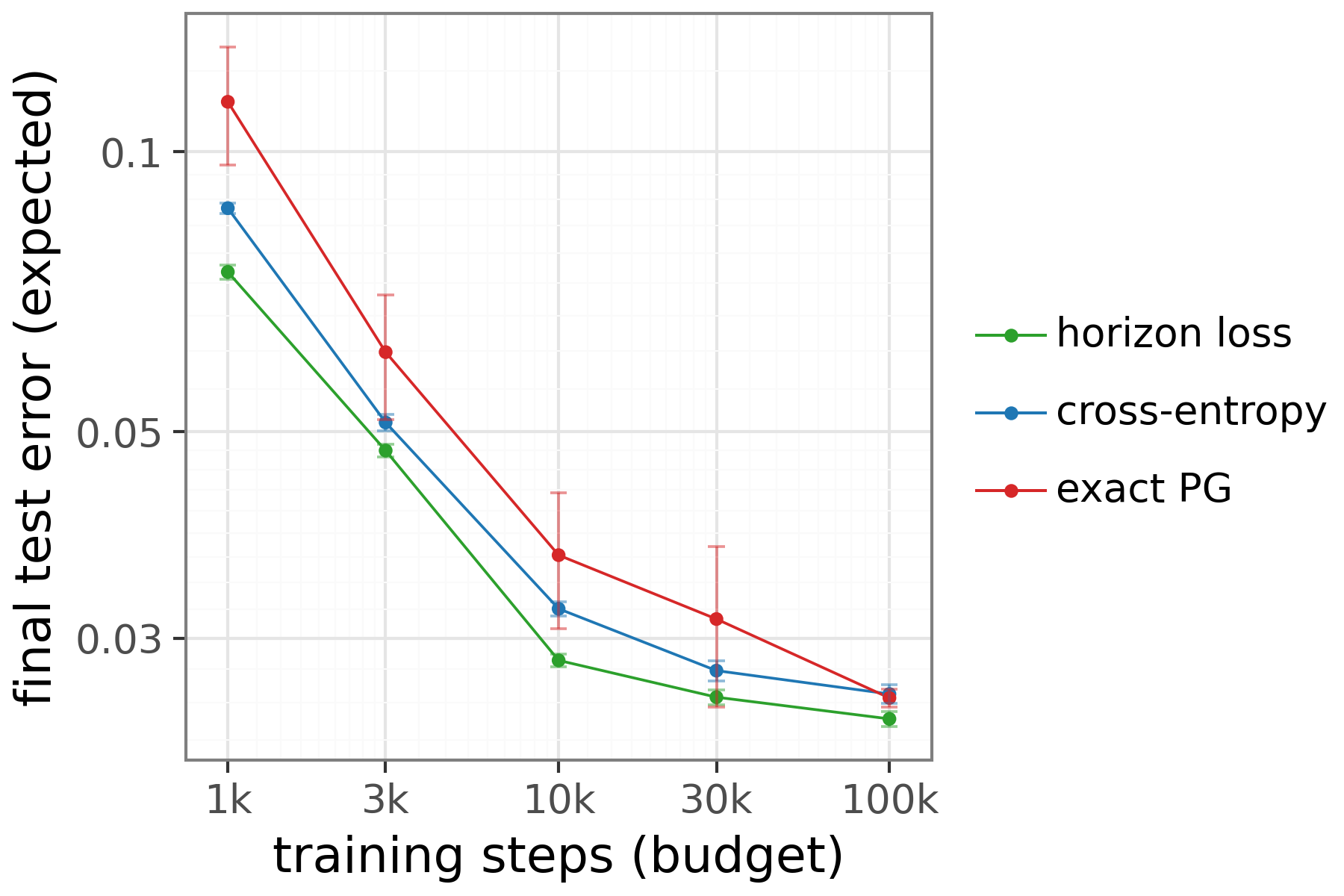}\caption{}\end{subfigure}
\caption{MNIST. (a) Final error against width. (b) Relative error reduction against width. (c) Final error against training budget.}
\label{fig:width}
\end{figure}

On MNIST, the expected-error gain holds at every width, and the top-1 gain grows with width from 0 to 10\% (Figure~\ref{fig:width}a,b).
At $\kappa = 1$, the horizon loss wins at every budget from 1k to 100k steps (Figure~\ref{fig:width}c).
Exact PG catches cross-entropy at 100k steps, so on MNIST its deficit is speed, not reachability.

\subsection{Horizon Cross-Entropy}
\label{app:hxent}

Applying the planning operator~\eqref{eq:planning} to cross-entropy gives $P_H[\CE](z) = \int_0^H \CE(z+s)\,ds$, whose negative gradient is $C_H(z)\,\nabla z$.
We train on $P_H[\CE]/(1-e^{-H})$, implementing the update by multiplying the cross-entropy gradient by the detached factor $C_H(z)/\big((1-e^{-H})(1-p)\big)$, capped at 10 (Appendix~\ref{app:details}).
The normalization recovers cross-entropy as $H \to 0$.

\begin{figure}[ht!]
\centering
\includegraphics[width=0.85\textwidth]{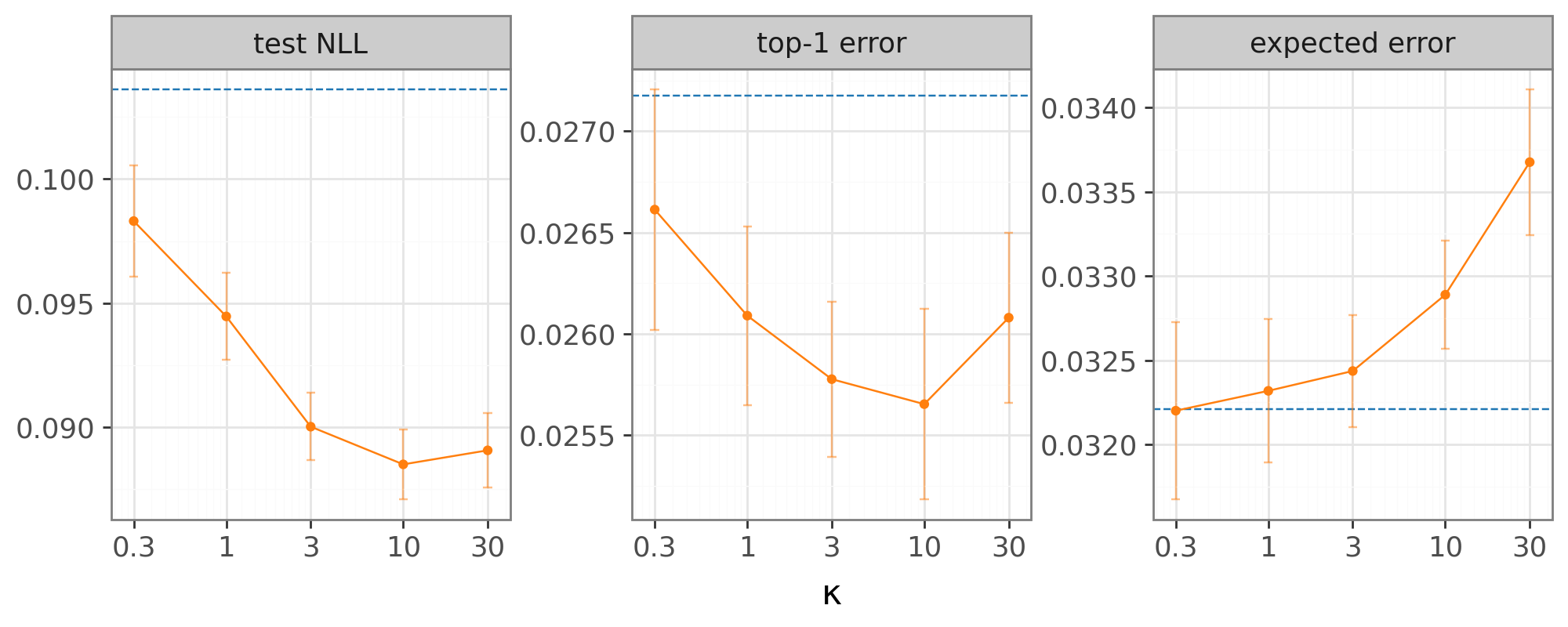}
\caption{Horizon cross-entropy against cross-entropy on MNIST: test NLL, top-1 error and expected error against $\kappa$.}
\label{fig:hxent_kappa}
\end{figure}

On MNIST, horizon cross-entropy has lower NLL and top-1 error than cross-entropy at every $\kappa$, and higher expected error for $\kappa \ge 1$ (Figure~\ref{fig:hxent_kappa}).
On ResNet-50 at a flat learning rate, NLL falls from 1.467 to 1.364 while expected error rises from 0.378 to 0.393.
Planning bounded accuracy eventually drops examples outside the remaining horizon; planning unbounded NLL keeps paying for examples with large reducible log-loss.

The NLL gain shrinks under cosine decay: $1.270 \pm 0.003$ ($n=3$) for horizon cross-entropy against $1.289 \pm 0.001$ ($n=10$) for cross-entropy, using the 10 headline runs without weight decay.
For reference, the horizon loss has test NLL 0.122 on MNIST and 1.77 on ResNet-50, against 0.103 and 1.467 for cross-entropy.

\section{Further Related Work}
\label{app:related}

\paragraph{Reward and likelihood.}
Sequence-level training directly optimizes expected task reward~\citep{ranzato2016sequence,shen2016minimum}, while reward-augmented maximum likelihood keeps the form of likelihood training~\citep{norouzi2016reward}.
Control as inference goes the other way, writing reward maximization as likelihood under an auxiliary optimality variable~\citep{levine2018reinforcement}.

\paragraph{RL post-training.}
DAPO filters sampled groups whose responses are all correct or all incorrect~\citep{yu2025dapo}, and Dr.~GRPO shows that question-level reward normalization itself reweights prompts by difficulty~\citep{liu2025understanding}.

\paragraph{Surrogate losses.}
Surrogate-loss theory~\citep{zhang2004statistical,bartlett2006convexity} starts from the difficulty of optimizing the discontinuous 0--1 loss and asks which tractable surrogate recovers the same Bayes decision.
Proper scoring rules~\citep{gneiting2007strictly} ask the same of probabilities.
Expected accuracy removes that obstacle, since its gradient is exact, and cross-entropy still learns faster.

\paragraph{Horizons in meta-learning and scheduling.}
Meta-gradient methods optimize hyperparameters or update rules by differentiating through training~\citep{andrychowicz2016learning,xu2018meta}.
Truncating the unroll can favor hyperparameters that look good over the short meta-objective but are poor for the full run~\citep{wu2018understanding}.
The horizon loss writes the horizon into the objective in closed form, with no unroll.
Budgeted training adapts the learning-rate schedule to a fixed training budget~\citep{li2020budgeted}, whereas here the remaining budget sets the loss.

\paragraph{Delight.}
Delightful Policy Gradient gates each sampled term by delight, the product of advantage and surprisal~\citep{osband2026delightfulpolicygradient,osband2026distributedpolicygradient}.
Across contexts, this moves the expected gradient toward the cross-entropy oracle, and the effect persists with infinite samples.
With zero baseline and unit temperature, its full-information expected update is $\tfrac{p}{1+p}\nabla_\theta\log p = -\tfrac12\nabla_\theta L_{\log 2}$ (Appendix~\ref{app:family}).
Its companion~\citep{osband2026doesgradientsparkjoy} pays for a backward pass only when delight exceeds a compute price, and so allocates compute rather than learning progress.

\paragraph{Label noise and example weighting.}
At $H = 0$, $\|\pi - e_y\|_1 = 2(1-p_y)$, and \citet{zhang2018generalized} find that this loss trains poorly.
Overparameterized networks memorize random and corrupted labels~\citep{zhang2016understanding,arpit2017closer}.
For every finite $H$, $L_H$ is bounded, and its range contracts as the horizon shrinks.
Self-paced learning~\citep{kumar2010self} schedules which examples enter training, and meta-reweighting learns example weights from a clean validation objective~\citep{ren2018learning}.
None derives its weights from the remaining optimization budget.

\paragraph{Softmax policy gradients.}
Work on softmax policy gradients studies a different failure, where the softmax geometry itself causes slow convergence or sensitivity to initialization even with exact gradients~\citep{mei2020global,mei2020gravitational}.
We retain the sigmoid learning curve but remove parameter sharing, which exposes how its local slopes allocate a finite budget across examples.

\end{document}